\documentclass[fleqn,10pt]{wlscirep}
\usepackage{xr}

\usepackage[utf8]{inputenc}
\usepackage[T1]{fontenc}
\usepackage{comment}
\includecomment{TTDnotes}

\usepackage{amsmath}
\usepackage{amssymb}
\usepackage{mathtools}

\usepackage{textcomp}
\usepackage{diagbox}
\usepackage{subcaption}
\usepackage{caption}
\usepackage{multirow}
\usepackage{booktabs}
\usepackage{colortbl}
\usepackage{footnote}
\usepackage{footmisc}
\usepackage{afterpage}
\usepackage{lineno}

\newcommand*\patchAmsMathEnvironmentForLineno[1]{%
  \expandafter\let\csname old#1\expandafter\endcsname\csname #1\endcsname
  \expandafter\let\csname oldend#1\expandafter\endcsname\csname end#1\endcsname
  \renewenvironment{#1}%
     {\linenomath\csname old#1\endcsname}%
     {\csname oldend#1\endcsname\endlinenomath}}%
\newcommand*\patchBothAmsMathEnvironmentsForLineno[1]{%
  \patchAmsMathEnvironmentForLineno{#1}%
  \patchAmsMathEnvironmentForLineno{#1*}}%
\AtBeginDocument{%
\patchBothAmsMathEnvironmentsForLineno{equation}%
\patchBothAmsMathEnvironmentsForLineno{align}%
\patchBothAmsMathEnvironmentsForLineno{flalign}%
\patchBothAmsMathEnvironmentsForLineno{alignat}%
\patchBothAmsMathEnvironmentsForLineno{gather}%
\patchBothAmsMathEnvironmentsForLineno{multline}%
}

\newcommand{\tauI}{\ensuremath{\tau_{I}}}
\newcommand{\tauV}{\ensuremath{\tau_{V}}}
\newcommand{\tauadap}{\ensuremath{\tau_{\text{AHP}}}}

\newcommand{\taureadout}{\ensuremath{\tau_{\text{readout}}}}

\newcommand{\robrace}[1]{\ensuremath{\left(#1\right)}}
\newcommand{\sqbrace}[1]{\ensuremath{\left[#1\right]}}

\newcommand{\ipsc}{\ensuremath{i_\text{PSC}}}
\newcommand{\iAHP}{\ensuremath{i_\text{AHP}}}

\newcommand{\ipscj}{\ensuremath{i_{\text{PSC}, j}}}
\newcommand{\ipsci}{\ensuremath{i_{\text{PSC}, i}}}
\newcommand{\iAHPi}{\ensuremath{i_{\text{AHP}, i}}}
\newcommand{\iAHPj}{\ensuremath{i_{\text{AHP}, j}}}

\newcommand{\vPSC}{\ensuremath{V_\text{PSC}}}
\newcommand{\vAHP}{\ensuremath{V_\text{AHP}}}

\newcommand{\vPSCj}{\ensuremath{V_{\text{PSC}, j}}}
\newcommand{\vAHPj}{\ensuremath{V_{\text{AHP}, j}}}

\newcommand{\tinp}{\ensuremath{T_\text{inp}}}
\newcommand{\tsim}{\ensuremath{T_\text{sim}}}
\newcommand{\Treadout}{\ensuremath{T_\text{readout}}}
\newcommand{\Tword}{\ensuremath{T_\text{word}}}

\newcommand{\deltat}{\ensuremath{\Delta t}}
\newcommand{\ms}{\ensuremath{\text{ms}}}

\newcommand{\dL}[1]{\ensuremath{\frac{dL}{d {#1}}}}
\newcommand{\partd}[2]{\ensuremath{\frac{\partial {#1}}{\partial {#2}}} }

\DeclareMathOperator{\mean}{mean}

\newsavebox\mytabularbox
\newcount\figwidthc
\newcount\textwidthc

\newcommand\totextwidth[1]{%
  \sbox{\mytabularbox}{#1}%
  \figwidthc=\wd\mytabularbox

  \textwidthc=\textwidth%
  \FPdiv\scaleratio{\the\textwidthc}{\the\figwidthc}%
  \FPmin\scaleratio{\scaleratio}{1}%
  \scalebox{\scaleratio}{\usebox{\mytabularbox}}%
}

\makeatletter
\newcommand{\mathleft}{\@fleqntrue\@mathmargin0pt}
\newcommand{\mathcenter}{\@fleqnfalse}
\makeatother

\title{A Long Short-Term Memory for AI Applications in Spike-based Neuromorphic Hardware}

\author[1,+]{Arjun Rao}
\author[1,3,+]{Philipp Plank}
\author[2]{Andreas Wild}
\author[1]{Wolfgang Maass*}
\affil[1]{Institute of Theoretical Computer Science, Graz University of Technology,
Inffeldgasse 16b, Graz, Austria}
\affil[2]{Intel Labs, Intel Corporation, 2111 NE 25th Ave, Hillsboro, OR 97124, USA}
\affil[3]{Intel Labs, Intel Corporation, Lilienthalstr. 15, 85579 Neubiberg, Germany}

\affil[*]{Corresponding author maass@igi.tugraz.at}
\affil[+]{These authors contributed equally to this work}

\keywords{neuromorphic, spiking, LSTM, AI, reasoning, Intel Loihi}

\begin{abstract}
Spike-based neuromorphic hardware holds the promise to provide more energy efficient implementations of Deep Neural Networks (DNNs) than standard hardware such as GPUs. But this requires to understand how DNNs can be emulated in an event-based sparse firing regime, since otherwise the energy-advantage gets lost. In particular, DNNs that solve sequence processing tasks typically employ Long Short-Term Memory (LSTM) units that are hard to emulate with few spikes. We show that a facet of many biological neurons, slow after-hyperpolarizing (AHP) currents after each spike, provides an efficient solution. AHP-currents can easily be implemented in neuromorphic hardware that supports multi-compartment neuron models, such as Intel's Loihi chip. Filter approximation theory explains why AHP-neurons can emulate the function of LSTM units. This yields a highly energy-efficient approach to time series classification. Furthermore it provides the basis for implementing with very sparse firing an important class of large DNNs that extract relations between words and sentences in a text in order to answer questions about the text.

\end{abstract}

\begin{document}

\flushbottom
\maketitle

\thispagestyle{empty}

\label{sec:introduction}

\begin{figure}[!htb]
    \centering
    \begin{minipage}{\textwidth}
        \includegraphics[width=\textwidth]{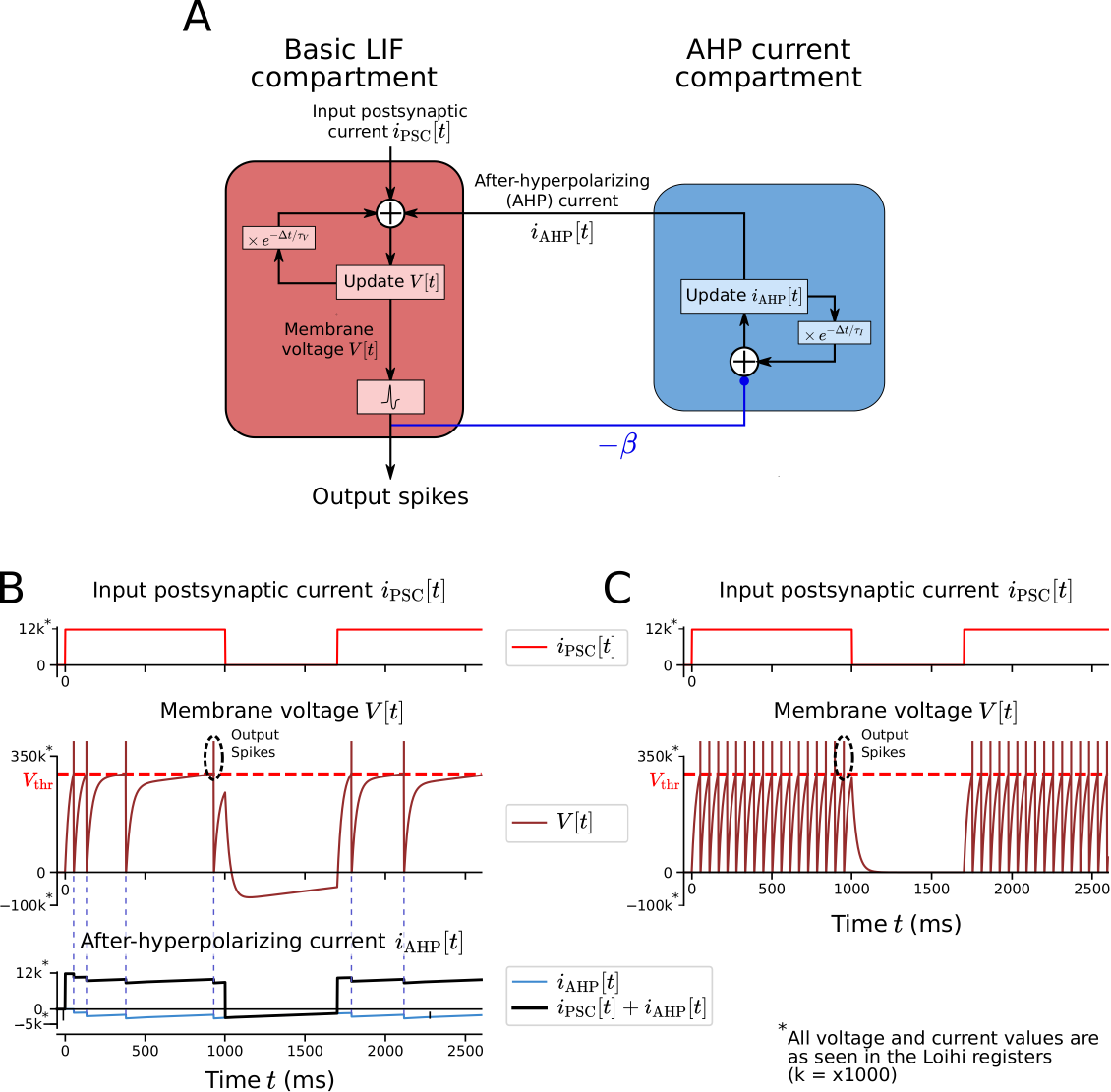}
        \renewcommand{\thempfootnote}{\arabic{footnote}}
        \caption{\textbf{Schematics and dynamics of a two-compartment AHP-neuron.}~\textbf{A)} Schema of the two-compartment neuron model. The second compartment adds a very slowly decaying AHP-current for each spike of the neuron. The AHP-current ($\iAHP$) becomes more negative by an amount $\beta$ whenever the neuron spikes, i.e., z(t) = 1, and decays exponentially between spikes with a large time constant $\tauadap$. This current, along with the input current $\ipsc$ are integrated to calculate the membrane voltage. This can be easily implemented on Loihi. The postsynaptic current (PSC) $\ipsc[t]$ caused by external input to the neuron and the AHP-current are leaky-integrated into the membrane voltage $V[t]$.
                 Spikes are emitted and $V[t]$ is reset each time the voltage crosses a firing threshold $V_{\text{thr}}$.
                 \textbf{B)} Response of an AHP-neuron to a sequence of two current steps as input. One sees how the AHP-currents (blue trace) add up, and reduce firing. Note the difference in its responses to the two input current steps. It shows a clear memory effect.
                 ~\textbf{C)} Response of the standard LIF-neuron model to the same input currents. Here the second compartment for AHP-currents in A has been removed. One sees that this neuron fires at a much higher constant spike-rate, and responds to the second current step in exactly the same way as to the first one.
                 }
        \label{fig:two-comp-model}
    \end{minipage}
\end{figure}

\afterpage{\clearpage}

Energy consumption is a major impediment for more wide-spread applications of new AI-methods that use DNNs, especially in edge devices. Spike-based neuromorphic hardware is one direction that promises to alleviate this problem. This research direction is partially inspired by the brain  that runs even more complex and larger neural networks with a total energy consumption of just 20W. A key factor of this astounding energy efficiency is that neurons in the brain emit a signal (spike) on average just a few times per second. In contrast, the units of a typical DNN emit an output value, and hence consume energy, by several orders of magnitudes more frequently. But it has remained an open problem which types of DNNs for modern AI solutions can be implemented in an energy-efficient manner with sparsely active neurons in neuromorphic hardware, see Fig. 9 of Davies et al.~\cite{davies2021}. In most cases this requires a rethinking of DNN design principles.

A more specific open problem is how the LSTM units of DNNs for sequence processing tasks can be implemented energy-efficiently in spike-based neuromorphic hardware. We show that a characteristic feature of biological neurons, the presence of slowly changing internal currents that have so far not been included in neuromorphic hardware models, endow networks of spiking neurons (SNNs) with similar working memory capabilities as LSTM units in DNNs. In particular, slow after-hyperpolarizing (AHP) currents reduce the readiness of biological neurons to fire again after recent firing activity. This effect is called spike-frequency adaptation in neuroscience, see Benda et al.~\cite{Benda2003} and Gutkin et al.~\cite{Gutkin2014}. Experimental data from the Allen Institute~\cite{allen2021} show that a fair number of neurons in the neocortex, e.g. over a third of excitatory neurons in the human frontal lobe, exhibit spike frequency adaptation. We show that AHP-neurons not only save energy by reducing firing activity, they also provide a principled alternative to LSTM units for solving sequence processing tasks, and they support training through BPTT (backpropagation through time).

Another major difference between biological neurons and standard spiking neuron models is that biological neurons keep their membrane potential in a rather narrow regime. In contrast, the membrane potential of models often assumes extremely negative values when the network is trained with regularization terms in order to induce low firing rates. This practically removes many of them from the current network computation. We introduce a new membrane voltage regularization principle that alleviates this problem, and supports the design of extremely sparsely firing spiking DNNs. 

We analyze functional implications of these two principles on a commonly used spike-based chip: Intel's neuromorphic chip Loihi~\cite{davies2018}, and find a significant reduction in the energy-delay product (EDP). In contrast to power, EDP accounts for the true energy and time cost per task/workload/computation. Simultaneously, these implementations demonstrate that two hallmarks of cognitive computations, both in brains and in machine intelligence, working memory and reasoning about relations between concepts or objects, can in fact be implemented more efficiently in spike-based neuromorphic hardware than in GPUs, the standard computing hardware for implementing DNNs.

\section*{Long short-term memory for spike-based neuromorphic hardware}
\label{sec:results}

\afterpage{\clearpage}

The value of a continuous variable can be stored in the memory cell of an LSTM, to which read- and write-access is controlled by auxiliary neurons with sigmoidal activation function. A straightforward implementation of an LSTM unit~\cite{Hochreiter1997} with spiking neurons arrives at rate-coding of the stored variable and the continuous-valued outputs of the auxiliary gates. Hence this cannot reach an 
energy-efficient sparse firing regime. Shrestha et al.~ \cite{shrestha2017spike} have shown that if one adds some clever sub-circuits and a store-and-release mechanism to the already rather complex circuitry of an LSTM unit, one can emulate it on IBM's TrueNorth chip~\cite{truenorth2015}, using substantially less energy than a GPU but causing larger latency. They tested computational performance on sequences up to length 36 (compare with sequence length 784 in the standard version of the sMNIST task that will be discussed in the next section).
More recently, Rezaabad et al.~\cite{Rezaabad2020} considered a spike-based implementation of LSTM units where each output of their sigmoidal gates is rounded to "spike" or "no-spike". They applied the resulting SNN to a simpler version of sMNIST where each handwritten digit was transformed into a time series of length 28 (presenting a whole row of pixel values at each time step). No implementation in neuromorphic hardware, hence no energy consumption analysis was given for this approach. 

Brains appear to use instead of LSTM-like units a different method for solving temporal computing tasks.  We propose that brains use instead slower dynamic processes of neurons, such as AHP-currents, see Fig.~\ref{fig:two-comp-model} A and B. This mechanism does not enable storage and maintenance of a continuous-valued variable like in the memory cell of a LSTM unit. But it supports an alternative strategy for computing with long short-term memory that is motivated by a
theoretical result about filter approximations, 
based on the Stone-Weierstrass theorem from mathematics (see Theorem 1 in Maass et al.~\cite{maass2002} for a suitable formulation). 
This theory implies that in order to classify time series, one does not have to store their earlier segments in memory cells. Rather, it suffices to satisfy the Pointwise Separation PRoperty (PSPR). We will explain the PSPR in the next section in the context of an application to a concrete time series classification task, and show that neurons with AHP-currents significantly improve the PSPR. 

AHP-currents can be realized easily by a two-compartment version of the standard leaky integrate-and-fire (LIF) spiking neuron model, see Fig.~\ref{fig:two-comp-model} A, and hence by any neuromorphic hardware such as Loihi~\cite{davies2018} or SpiNNaker~\cite{Furber2014} that supports multi-compartment neurons. When the neuron fires a few spikes within a short time window, the resulting AHP-currents add up, see Fig.~\ref{fig:two-comp-model} B, and delay or halt further firing. Their impact on firing outlasts the gap of 700ms between the two input current steps if $\tauadap$ is sufficiently large, and reduces the firing response already at the beginning of the second input current step. In this way information can be stored in the hidden variable $\iAHP[t]$ for longer time spans, and can be communicated to other neurons in the network or to network readouts when stimulated by an input current.

AHP-currents being inhibitory reduce the firing activity of a neuron: Compare for example the firing response of neurons with and without AHP-currents in Fig.~\ref{fig:two-comp-model} B and C. Neurons with AHP-currents can also be used for many standard network computations, and thus one can view SNNs which include them as in-memory-computing architectures, since they do not require energy-consuming shuffling of information between processors and working memory units. We refer to neurons with AHP-currents as AHP-neurons and to neurons without AHP-currents as LIF-neurons. Networks that contain AHP-neurons are termed AHP-SNNs.

\begin{figure}[b!]
    \centering
    \includegraphics[width=\textwidth]{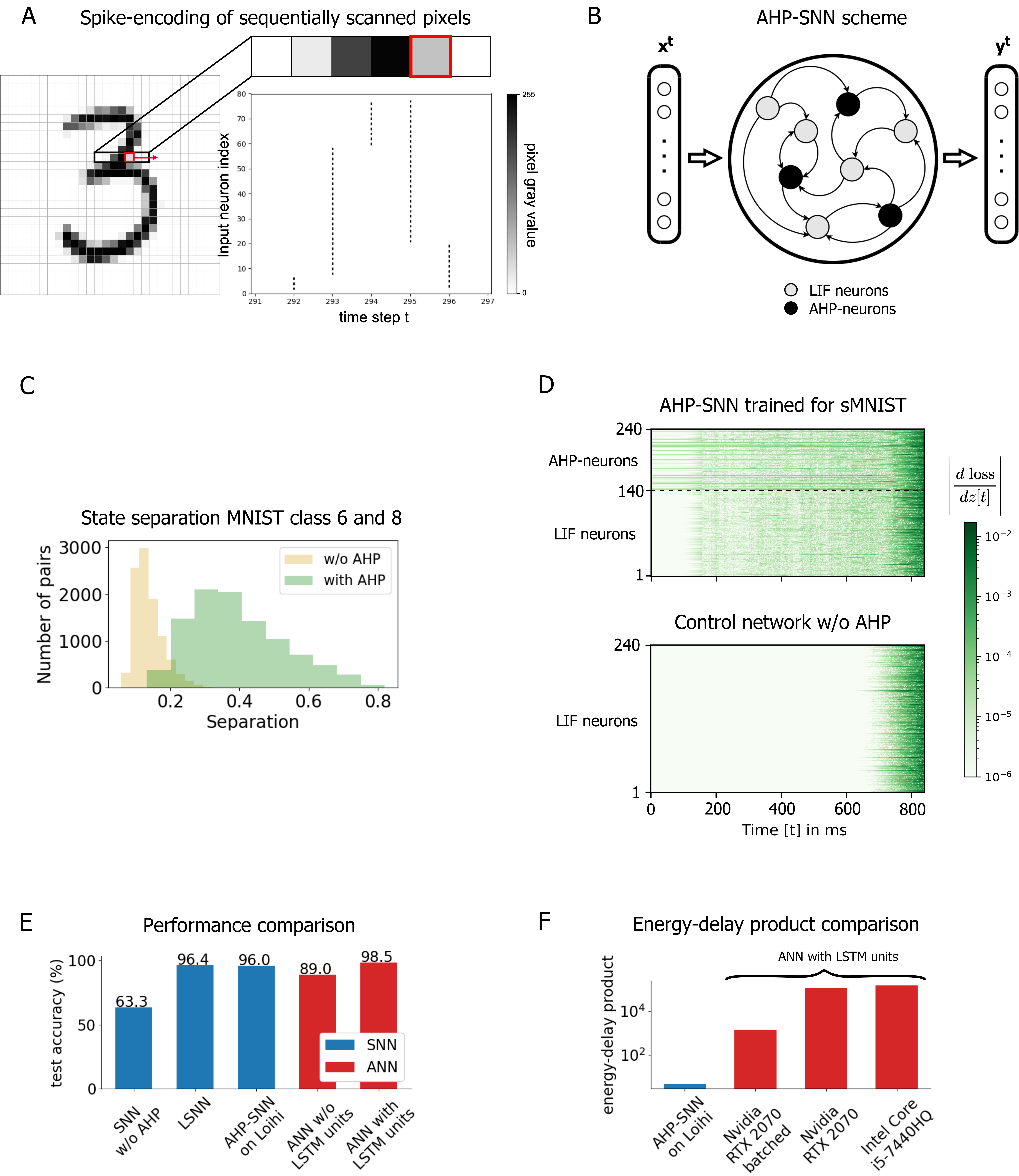}
    \caption{\textbf{Illustration of the sMNIST task, pointwise separation property PSPR and gradient transmission in AHP-SNN networks, comparisons of performance and energy consumption for various network types.}
        \textbf{A)} The input pixels are fed in sequentially, row by row, encoded by spikes based on a threshold crossing method. 40 thresholds were used, represented by 80 input neurons, which send spikes depending on the change of the pixel value with respect to the previous one.~\textbf{B)} The AHP-SNN  consists of an input layer sending spikes, a recurrently connected network of LIF- and AHP-neurons, and a linear readout layer.~\textbf{C)} PSPR for SNNs with (green) and without (yellow) AHP-neurons consisting of 240 fully connected neurons with random weights: histogram of distances of normalized network states for input pairs 6 and 8. (continuation on next page)
        } 
    \label{fig:sMNIST}
\end{figure}

Since AHP-currents change only on a slow time scale, the gradients of BPTT can pass through the associated hidden variable without loss of precision, see Methods. Furthermore, the gradient can go  repeatedly through these variables when it moves backwards in time, without being subject to exponential growth or decay, see Fig.~\ref{fig:sMNIST} D for a demonstration. In this way, gradients propagate substantially better even to earlier time slices of the neurons without AHP-currents, since they can use the slowly changing AHP-currents of other neurons as intermediate stepping stones. Therefore AHP-SNNs can in general be trained more effectively through BPTT than SNNs without AHP-neurons. LSNNs, introduced by Bellec et al.~\cite{bellec2018}, provide similar advantages, but employ neurons with a time-varying firing threshold that cannot be efficiently implemented on Loihi.

\addtocounter{figure}{-1}
\afterpage{\begin{figure} [t!]
    \caption{\textbf{D)} Backpropagation of the gradient of loss w.r.t neuron spike output $z[t]$ through time for a sample image as input, for SNNs with (top) and without AHP-neurons (bottom). AHP-currents sprovide "highways back into the past" like memory cells of LSTM units, that also improve gradient transmission to neurons without AHP-currents. ~\textbf{E)} Classification accuracy for sMNIST of the AHP-SNN network on Loihi in comparison to SNNs without AHP-neurons; spiking neurons with adapting thresholds (LSNN), and non-spiking ANNs consisting of sigmoidal neurons and LSTM units was compared to the full precision LSNN; comparison results are from Bellec et al.~\cite{bellec2018}~\textbf{F)} Comparison of the  EDP for the AHP-SNN on Loihi and for a network of LSTM units of the same size on standard GPUs, utilizing parallel evaluation of 100 samples at the same time (batched) and on a standard CPU. }
\end{figure}}

\section*{Comparing the energy consumption of SNNs and non-spiking RNNs on a standard time-series classification benchmark task}
\label{sec:smnist}

    In order to test the energy efficiency of the proposed replacement of LSTM units via AHP-SNNs, we use a classical time series classification task: sequential MNIST (sMNIST). Here the 784 pixel values of each 28 x 28 image of a handwritten digit from the MNIST dataset~\cite{lecun2010mnist} are presented sequentially, one pixel at each ms, in a fixed order: row by row, starting with the top row, see the left part of Fig.~\ref{fig:sMNIST} A.
    The gray values of pixels are encoded by spikes through a population of spiking input neurons that fire when the gray value crosses some threshold, where each neuron in the population relates to specific thresholds (see bottom right of Fig.~\ref{fig:sMNIST} A). The computational task is to classify the underlying digit from the time series of the pixel values that are presented sequentially during 784 ms. The network output is encoded by that one of the 10 readout neurons that has the highest membrane potential at time 840 ms, i.e., 56ms after the pixel sequence has been provided. An additional input neuron fires during these 56 ms. This time series classification task is obviously more difficult than the standard version of MNIST~\cite{lecun2010mnist} with batch inputs, since information from pixels in the upper pixel rows needs to be remembered and combined with information from pixels in lower rows. This difficulty is for example obvious for the case of digits 6 and 8, where the most salient differences are contained in the earliest part of the resulting input time series.
    
   In order to solve this time series classification task, a recurrent network needs to have after $T = 784\ms$, when the input sequence finishes, different network states for input sequences that represent different digits: Otherwise it could not activate different readouts for these input sequences. The cited theoretical result on filter approximations states that this necessary condition is under some constraints also sufficient. In other words, a neural network should aim at satisfying the PSPR. The PSPR demands in this case that the network states at time $T$ should be different for any pair of input time series that require different network outputs, in particular for any pair of handwritten digits 6 and 8. Since the readouts from the recurrent network need to work in the presence of noise, which arises here from different writing styles of the same digit, one should try to satisfy a stronger version of the PSPR: that the network states at time $T$ are not only different but significantly different. 
We show in Fig.~\ref{fig:sMNIST} C the histogram of pairwise euclidean distances between normalized network states (low-pass filtered spike trains of all network neurons) at time $T = 840\ms$ for pairs of digits 6 and 8. This is shown for two variants of the network of Fig.~\ref{fig:sMNIST} B, using in both cases random weights for recurrent connections, rather than trained weights: One containing AHP-neurons, and one without AHP-neurons. One clearly sees that AHP-neurons substantially enhance the separation of network states. This improved separation property via AHP-neurons is reflected in the time series classification capability of these untrained networks: With AHP-neurons, trained linear readouts from the considered network states at time $T$ achieve an accuracy of 89.8\% for the classification of digits 6 and 8, but only 69.6\% without AHP-neurons.

Another important property of LSTM units is that they can be trained through BPTT without gradients exploding or vanishing when they are backpropagated in time. In Fig.~\ref{fig:sMNIST} D, we show the gradient of the loss propagating through an AHP-SNN network fully trained on the sMNIST task (top), versus a randomly initialized control network without AHP-neurons (bottom). We demonstrate here that the AHP-currents of multi-compartment neurons provide a corresponding "highway into the past" for gradient propagation because these currents change only slowly.

  We examined solutions of this time series classification task by various types of recurrent neural networks, in particular for AHP-SNNs that contained both AHP-neurons and LIF-neurons without AHP (see Fig.~\ref{fig:sMNIST} B). Fig.~\ref{fig:sMNIST} E shows the performance of an AHP-SNN consisting of 240 neurons, 100 of which were AHP-neurons. In order to arrive at a recurrent network that can be efficiently implemented on Loihi in spite of its limitation on the number of synaptic connections (see Methods), we considered AHP-SNNs with just 20\% connectivity. In order to achieve nevertheless high performance we used BPTT and DEEP-R~\cite{bellec2018deep} for training where synaptic connections are occasionally removed and randomly added in order to arrive at a well-functioning network architecture. This adaptive rewiring strategy is inspired by the ongoing removal of weak synaptic connections (spines) and the creation of new ones in biological neural networks. We show for comparison also performance results for several other spiking and non-spiking neural networks on the same task. In spite of the fact that weight resolution was limited to 8 bits in the Loihi implementation of the AHP-SNN, its performance was on the same high level as other spiking and non-spiking networks that were not subject to this constraint. The comparison between the performance of the SNNs with and without AHP-neurons shows that the high performance level of AHP-SNNs is only to a minor extent due to their common recurrent SNN architecture.  Details on the architecture and parameters of all networks can be found in Methods. 

    In order to compare accuracy, latency and energy consumption of LSTM networks on conventional hardware with that of the AHP-SNN on Loihi, we measured the EDP of an LSTM network~\cite{Hochreiter1997} for the same task on CPUs and GPUs. 
    The energy-delay product (EDP) of the AHP-SNN running on Loihi is by four  orders of magnitude lower than that of the traditional LSTM architecture on CPU or GPU (see Fig.~\ref{fig:sMNIST} F) for batch size 1, with Loihi outperforming over 2x on latency and over a 1000x on energy consumption per inference (see Table~\ref{tab:power-results}). Details regarding the benchmark procedure can be found in the Methods and Supplement.

    A reason for this significant improvement in latency and energy consumption compared to LSTMs on conventional hardware is the relatively small network size that suffices for solving this task.
    Thus, the network fits on a single chip of Loihi and uses only one neuro-core.
    The neuro-core is the fundamental computational unit of Loihi that has a shared SRAM.
    Thus keeping spike traffic within a neuro-core is the fastest and most efficient way to process spikes with Loihi.
    Performance in conventional neural networks is typically increased by using parallel processing of batches of data to increase the throughput.
    Even with a batch size of 100 on the GPU the spiking network on Loihi operating in the batch size 1 regime is still more energy efficient.
   
\section*{Design of a very sparsely active large DNN for energy-efficient spike-based relational reasoning}
\label{sec:relnet}

    We wondered whether this implementation of working memory in spiking neurons could be used to also implement large DNNs for more demanding sequence processing tasks in an energy-efficient manner in spike-based neuromorphic hardware. Therefore we implemented and tested a spiking variant of the relational network (RelNet) of Santoro et al.~\cite{santoro2017} on Loihi, which we refer to as the Spiking RelNet. The question of whether this can be done in an energy-efficient manner is quite non-trivial since the Spiking RelNet consists primarily of feed-forward networks. The Spiking RelNet takes as input a set of $K$ objects (sentences) and a single question that are encoded respectively by input spike trains $o_1(t),\ldots,o_K(t)$ and $q(t)$, see Fig.~\ref{fig:relational_network} B. As indicated in Fig.~\ref{fig:relational_network} A it computes the function
    \begin{equation}
        RN\robrace{\sqbrace{o_1(t), o_2(t), \ldots, o_K(t)}, q(t)} = f_\phi\robrace{f_{agg}\robrace{\sum_{1 \le i \le j \le N}{g_\theta\robrace{o_i(t), o_j(t), q(t)}}}} 
        \label{eqn:spiking-relnet-definition},
    \end{equation}
    with the output given through one-hot encoding of words by readout units. The only recurrent network modules are the ones indicated as module B of Fig.~\ref{fig:relational_network}, that transform each input sequence (a sentence of words in natural language) into spiking activity of 200 neurons within a compressed time span of 37ms. This input embedding of sentences was carried out by LSTM networks in the RelNet\cite{santoro2017}, and is carried out by AHP-SNNs in the Spiking RelNet.
    In the next processing step (see Fig.~\ref{fig:relational_network} C), the resulting compressed spike codes for each pair of sentences in the story and for the question are processed in parallel by a copy of a feed-forward LIF-neuron network that implements the relational function $g_\theta\robrace{o_i(t), o_j(t), q(t)}$ that extracts salient relational information for question $q$ from the two sentences.
    The outputs of these network modules are superimposed and connected to a LIF-neuron layer one-to-one, which implements the element-wise function $f_{agg}$ (see Fig.~\ref{fig:relational_network} D).
    The readout network $f_\phi$ processes the output of $f_{agg}$ through another feed-forward LIF-neuron network. The feed-forward networks don't use the AHP-current. The answer to the question $q$ is then given by an application of soft-max to one-hot readout neurons that each favor a particular word as the answer (see Fig.~\ref{fig:relational_network} E). For more details, see Methods and Supplement.

\begin{figure}[!t]
        \centering
        \includegraphics[width=\textwidth]{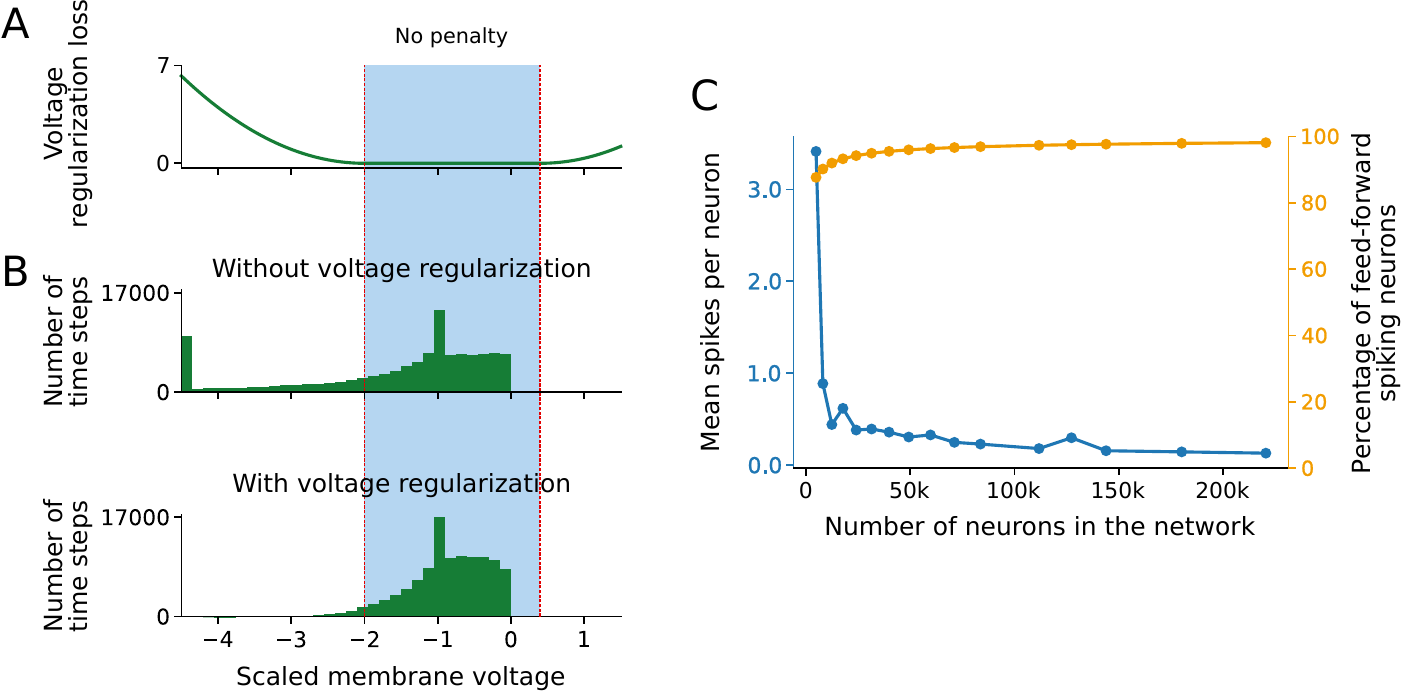}
        \caption{\textbf{Illustration of voltage regularization and its its capability to enforce --in conjunction with spike rate regularization-- a sparse firing regime.}~\textbf{A)}~The voltage regularization penalty as function of the value taken by the scaled membrane voltage at a particular time step. The scaled membrane voltage is as defined in Eq.~\ref{eqn:scaled-voltage-def}. A value of 0 corresponds to the spiking threshold, and a value of -1 corresponds to the value of the voltage corresponding to a zero input PSC. The membrane voltage is thus penalized if the scaled voltage is outside the range $\left[-2, 0.4\right]$. \textbf{B)}~The distribution of the scaled voltage values across different batches, neurons, and time steps with and without regularization. \textbf{C)}~The spikes used per neuron in relation to the network size (which varies for different story sizes). One observes that larger networks use fewer spikes per neuron as a result of spike rate regularization combined with voltage regularization, which results in savings in energy when run on hardware.}
        \label{fig:regularization-stats}
    \end{figure}

\begin{figure}[!htb]
        \centering
        \includegraphics[width=0.97\textwidth]{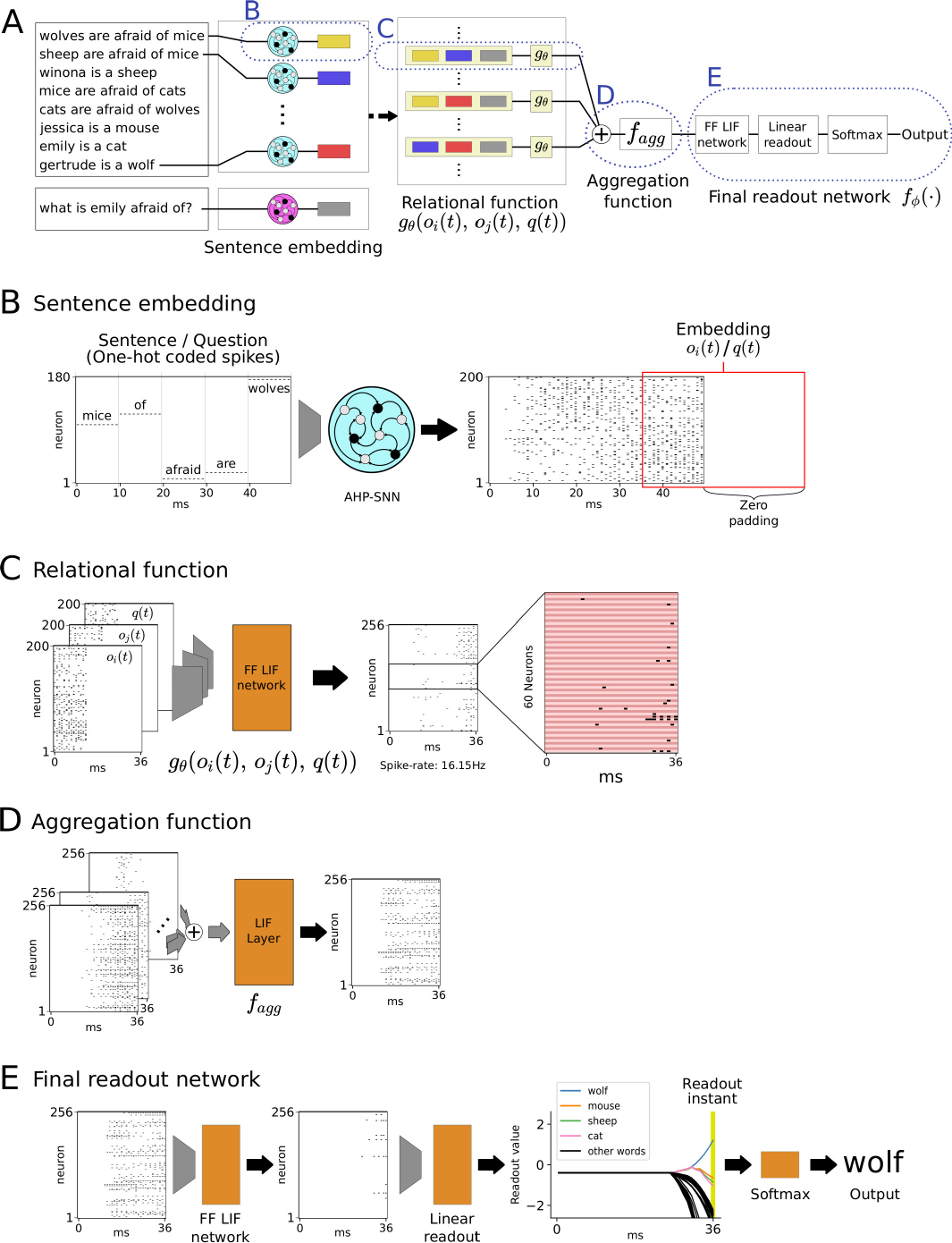}
        \caption{\textbf{Spiking RelNet implementation with very sparse firing.}~
            \textbf{A)}~The top-level Spiking RelNet architecture. We embed each sentence and the question into spike sequence objects $o_i(t)$ and $q(t)$ respectively via an AHP-SNN.
            For each pair of sentence objects $o_i, o_j\, : \, 1\le i\le j \le 20$, we apply the relational function $g_\theta$ to the triplet $(o_i(t), o_j(t), q(t))$.
            (continued next page)}
        \label{fig:relational_network}
    \end{figure}

    A large fraction of the Spiking RelNet is formed by feed-forward layers, especially as the number of sentences increases (see Fig.~\ref{fig:regularization-stats} B).
    This is problematic from the perspective of energy efficiency, since prior emulations of feed-forward networks in spike-based hardware suggested that their potential energy and latency advantage is encumbered by the need for relatively large firing rates, instead of event-based processing, in order to achieve good performance ~\cite{davies2021}.
    In contrast, our Spiking RelNet uses on average substantially less than one spike per neuron for the whole computation, see Fig.~\ref{fig:regularization-stats} C. We achieved this through an aggressive
    spike-rate regularization. But since this tends to move the membrane potential of many neurons into a strongly negative regime where they can no longer participate in the network computation, we complemented that by a new membrane voltage regularization, that keeps membrane voltages in a suitable range. In addition we moved the DNN into an event-based processing regime by forcing it to produce its network output at a particular time point, and by using neurons without refractory period and a rather short membrane time constant of 7ms (see Supplement for data on the impact of that).

    \afterpage{\clearpage}

    The hardware implementation required changes to the network architecture including restricting the ordered pairs to $1 \le i \le j \le N$ to allow the network to efficiently infer sentence ordering, as well as the introduction of $f_{agg}$ to tackle fan-in constraints.
    The network which solves the longest stories (20 sentences) contains 238604 neurons. The optimal implementation of this network on Loihi, conforming to connectivity constraints (see Methods, Supplement), requires 2308 neuro-cores spread across 22 chips (128 neuro-cores/chip, see Fig.~\ref{fig:board} B and C), which increases the energy consumption compared to the sMNIST case. Additionally, we demonstrate a method of placing the Feed-Forward layers across Loihi chips that minimizes congestion during spike transport (see Fig.~\ref{fig:board} D) which provides significant improvements of the EDP, see Fig.~\ref{fig:board} E (details in Methods and Supplement).
    Since the Spiking RelNet requires much less than 1 spike per neuron, it is likely to provide substantially larger energy-savings in future hardware which does not have this connectivity constraint.

    \afterpage{
    \addtocounter{figure}{-1}
    \begin{figure} [t!]
        \caption{The outputs of the relational function are aggregated in a LIF-neuron Layer $f_{agg}$ and then passed to the final readout function $f_\phi$.
            \textbf{B)}~The embedding scheme, where each word is provided for $\Tword = 10\ms$ with one-hot coded spikes, aligned so that the first word is provided at the very end of the duration.
            The spikes in the last $\tinp=14\ms$ are padded to a length of $\tsim=37\ms$ (red box) to form a time-compressed sentence embedding $o_i(t)$ and $q(t)$.
            \textbf{C)}~An instance of the spiking relational function $g_\theta$ operating on a sample triplet $(o_i(t), o_j(t), q(t))$. The blowup of the spike raster for 60 sample neurons makes apparent the resulting very sparse firing activity throughout the network (alternating light/dark shading demarcates the 60 neurons).
            \textbf{D)}~The aggregation layer is a layer of LIF-neurons that receive one-to-one connections from each relational function instance.
                        This aggregates the spike trains from across the relational function instances and outputs a spike sequence for the readout network.
            \textbf{E)}~The final readout function consists of a three layer feed-forward LIF-neuron network followed by a linear readout (with one neuron per word in the dictionary), that integrates synaptic inputs only during the last 10ms. 
                       The value of the readout at the final time step (marked as yellow bar) provides input to the softmax, whose output produces the the final answer through one-hot encoding of words.}
    \end{figure}
\begin{table}[ht!]
\centering
\bgroup
\def\arraystretch{1.5} 
\begin{tabular}{lccccccc}

\multicolumn{1}{c|}{}                                               & \multicolumn{2}{c|}{\textbf{sMNIST}}               & \multicolumn{5}{c}{\textbf{Relational reasoning}} \\
\multicolumn{1}{c|}{}                                               & GPU     & \multicolumn{1}{c|}{CPU}                           & \multicolumn{5}{c}{GPU}                           \\ \hline
\multicolumn{1}{l|}{\textbf{\# cores on Loihi}}                     & 1       & \multicolumn{1}{c|}{1}                             & 124      & 332      & 700     & 1552    & 2320    \\
\multicolumn{1}{l|}{\textbf{\# sentences (RR)}}                     & -       & \multicolumn{1}{c|}{-}                             & 2        & 6        & 10      & 16      & 20      \\ \hline
\multicolumn{1}{l|}{\textbf{Energy ratio}}                          & 7.467x  & \multicolumn{1}{c|}{4.774x}                        & 16.49x   & 11.92x   & 7.78x   & 5.32x   & 4.36x   \\
\rowcolor[HTML]{EFEFEF} 
\multicolumn{1}{l|}{\cellcolor[HTML]{EFEFEF}\textbf{Latency ratio}} & 2.82x   & \multicolumn{1}{c|}{\cellcolor[HTML]{EFEFEF}5.89x} & 0.73x    & 0.56x    & 0.44x   & 0.33x   & 0.38x   \\
\multicolumn{1}{l|}{\textbf{EDP ratio}}                             & 21.026x & \multicolumn{1}{c|}{28.134x}                       & 12.10x   & 6.73x    & 3.41x   & 1.73x   & 1.67x  \\
\\[-10pt]
\multicolumn{8}{l}{All ratios are shown against Loihi.} 
\end{tabular}
\egroup
\caption{\textbf{Benchmarking results.}~Comparison and scaling analysis of the spiking relational network on Loihi against the corresponding ANN on CPU and GPU (HW specifications in Supplement). For the scaling analysis of the Spiking RelNet the data set was grouped by number of sentences per sample which in turn determines the number of configured AHP-SNNs and therefore neuro-cores per sample. All measurements were done using 250 input samples, except for network size 16 where only 100 samples were used, as there are not enough test samples containing exactly 16 sentences. The energy per inference was calculated using total power values. More detailed results can be seen in the Supplement.}
\label{tab:power-results}
\end{table}

    }

 

    We tested the performance of the resulting spike-based RelNet implementation on Loihi for a standard benchmark dataset for question-answering; The bAbI dataset introduced by Weston et al.~\cite{weston2015}, that were also used for testing RelNet by Santoro et al.~\cite{santoro2017} This dataset consists of 20 different types of tasks, that each probe different challenges in reasoning about relational information contained in a set of sentences, i.e., a story. The questions are formulated in such a way that an answer can be given with a single word via one-hot encoding in the output (or with a sequence of two words in the case of path planning in Task 19; one has here an output category for each such possible sequence).
    According to the convention of Weston et al.~\cite{weston2015} and Santoro et al.~\cite{santoro2017} a task is considered as being solved if the network has an error rate of at most $5\%$ on instances of the task that had not been used for training.
    When applying a RelNet to solve this task, each sentence (question) forms an object $o_i$ ($q$) that is embedded via AHP-SNNs to a spiking representation $o_i(t)$ ($q(t)$), which are then processed as described above. The details pertaining to input encoding are discussed in the Methods.


    %
 
    We trained the Spiking RelNet on the combined data of 17/20 bAbI tasks, excluding those which the non-spiking RelNet wasn't able to solve.
    We observe that the Spiking RelNet can solve 16 out of the 17 tasks to under 5\% error. We show in the Supplement that networks using more time steps and larger time constants, with consequently greater energy consumption and latency, solve all 17 tasks.



\subsubsection*{Energy-efficiency of RelNet in spike-based neuromorphic hardware}

    We compared the energy consumption and delay of the spike-based implementation of RelNet on Loihi with GPU implementations of the ANN RelNet from Santoro et al.~\cite{santoro2017}, see Table~\ref{tab:power-results}. One sees that the spike-based implementation consumes between 4 and 16 times less energy than the GPU implementation. The energy savings are lower for longer story sizes, apparently because these require the use of substantially more Loihi chips, and inter-chip communication appears to be less energy efficient in this spike-based hardware. One should also note that the average length of a story for the 16 datasets that we consider is just 6.5 sentences. The computation time on Loihi was slightly larger than on the GPU. But nevertheless, the resulting EDP remained lower for Loihi. For the longest and therefore slowest story size the average computation time per sample is 6.54~ms wall-clock time, which would still be sufficient for online applications like voice control or virtual assistants.

    \afterpage{\clearpage}
    \begin{figure}[!htb]
        \centering
        \includegraphics[width=380pt]{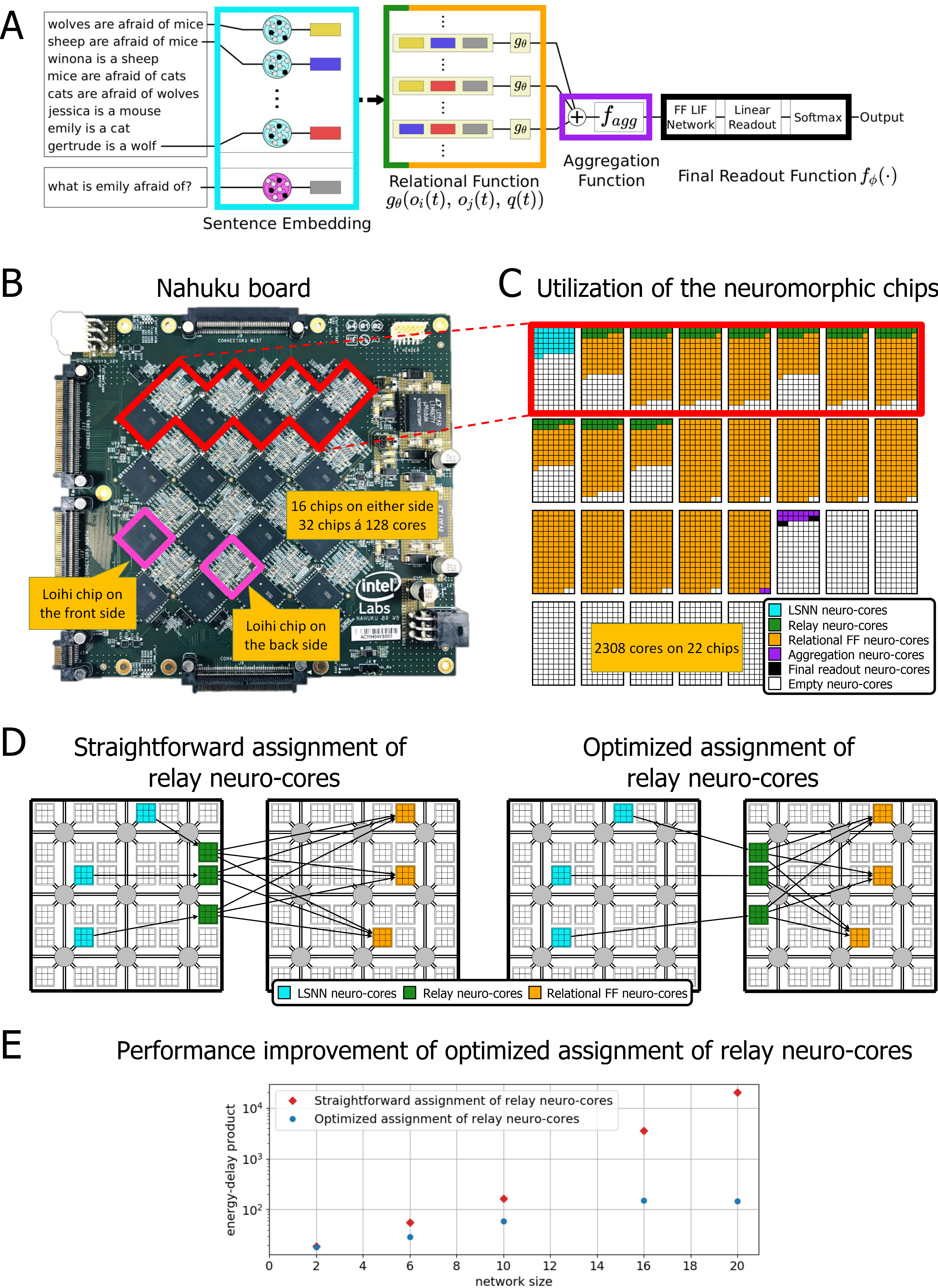}
        \caption{\textbf{Spiking RelNet placement and optimization on Loihi}~\textbf{A)}~Highlighting the different parts of the Spiking RelNet with the color code used in C and D. \textbf{B)}~The Spiking RelNet was configured on a Nahuku board with 32 Loihi chips. On each side of the board 16 chips are placed in a checkerboard pattern. Each chip has 128 neuromorphic computing units or neuro-cores.
            \textbf{C)}~The full scale Spiking RelNet which can solve tasks up to 20 sentences utilizes 2308 neuro-cores on 22 chips. The detailed mapping of the different layers is shown. In order to minimize cross-chip spikes some chips were not fully utilized. \textbf{D)}~Straightforward assignment of relay neuro-cores on the same chip as the source neuro-cores on the left side and optimized assignment of the relay neuro-cores on the chip of the target neuro-cores, to minimize inter-chip spike traffic, on the right side. Note that the target neuro-cores must be assigned carefully to do this efficiently.
            \textbf{E)}~Shows the benefit of the optimized assignment in energy-delay product as a function of network size as measured by number of bAbI sentences which roughly corresponds to the number of Loihi chips for the RelNet solving bAbI tasks on Loihi. Thus we see that the placement of the Spiking RelNet is optimized in terms of resource utilization as well as network performance on the hardware.} 
        \label{fig:board}
    \end{figure}

\section*{Discussion}

\label{sec:discussion}
We have shown that a key tool for sequence processing by DNNs in machine learning and AI, LSTM units, can be replaced in spike-based neuromorphic hardware by neurons with a biologically inspired mechanism for spike frequency adaptation (SFA); AHP-currents. This approach is supported by a theoretical principle, the pointwise separation property (PSPR) from filter approximation theory (see Fig.~\ref{fig:sMNIST} C). Adding a comportment for AHP-currents to the neuron model also has the advantage that it enhances the results of BPTT training by creating "highways back into the past" for the backwards propagation of gradients, see Fig.~\ref{fig:sMNIST} D. 
Since AHP-neurons can also be used for generic network computations, this solution does not require a separation of units for computing and working memory, it is an in-memory computing solution. This reduces latency and energy consumption that generally arise from traffic between computing and memory units. The resulting spike based solution for solving the benchmark time series classification task sMNIST turns out to be by three orders of magnitude more energy efficient than state-of-the-art implementations of LSTM networks on CPUs and GPUs, while achieving lower latency and virtually the same accuracy.

We have also shown that the use of AHP-neurons enables us to port really large DNNs that involve besides LSTM units large feed-forward network components into spike-based hardware. We have focused on the example of relational networks, since these enable a qualitative jump in AI capabilities by supporting reasoning about relationships between objects in a story or image. An energy efficient spike-based implementation of relational networks requires methods that enable them to use rare events (spikes) in time instead of rate-based neural codes, also in their feed-forward network modules. We have shown that an unprecedented sparseness of much less than one spike per neuron during the whole computation is achievable for relational networks, see Fig.~\ref{fig:regularization-stats} C. This became possible through a new voltage regularization method during training, in combination with an output convention that forced the network to produce its decision at a specific point in time, and a spiking neuron model with a short membrane time constant and no refractory period that especially supports computational operations that are strictly local in time. The resulting implementation of relational networks on Loihi provides in fact the first example of a large DNN om Loihi where this spike-based hardware becomes more energy efficient than GPUs, see Fig. 3 of Davies et al.~\cite{davies2021}. Furthermore the resulting very sparsely active Spiking RelNet is likely to become substantially more energy efficient on current and future neuromorphic hardware where the number of synaptic connections of the neurons in a neuro-core is less constrained than on Loihi. This constraint forced us to distribute the Spiking RelNet over 2308 neuro-cores on 22 Loihi chips, thereby increasing latency and energy consumption. But even in spite of that, a significant reduction of the EDP for relational networks compared with GPUs could be shown. In fact, the review of DNN implementations on Loihi~\cite{davies2021} arrives at the conclusion that it represents “the largest deep learning network to date showing gains compared to conventional architectures”. 
Thus we have shown that relational networks represent a class of DNNs that are more suitable than CNNs for porting them to spike-based hardware. Based on the results of Santoro et al.~\cite{santoro2017} one can expect that relational networks in neuromorphic hardware can be used not only for solving question-answering tasks in natural language, but also for reasoning about relations between objects in an image or in an auditory scene. This can provide a qualitative jump in AI capabilities of energy-efficient neuromorphic hardware.

Another interesting next step will be to enable on-chip training of these spike-based alternatives to LSTM networks by using e-prop instead of BPTT, which has already been shown to work very well for LSNNs (Bellec et al.~\cite{Bellec2020}). Also one-shot learning capability has been demonstrated for these spiking networks (Scherr et al.~\cite{Scherr2020}), and it is likely that the required method will also enable one-shot on-chip training of these networks.

Finally, adding AHP-currents to spiking neuron models on neuromorphic hardware can be seen as a first step towards porting more sophisticated point neuron models for neurons in the neocortex (Billeh et al.~\cite{Billeh2020}) into such hardware. If one adds one further current in an additional neuron compartment it will become possible to implement the diverse array of GLIF3 neuron models of the arguably most advanced model for generic cortical microcircuits from ~\cite{Billeh2020} in neuromorphic hardware. This will open the door to new uses of neuromorphic hardware in computational neuroscience: For simulating large state-of-the-art models for neural networks of the brain substantially faster and with less energy than currently possible. This has the potential of becoming a major new application of neuromorphic hardware such as Loihi or SpiNNaker~\cite{Furber2014} that supports the implementation of such biological refinements of the standard spiking neuron model. In particular, neuromorphic hardware may become in this way important for answering an important open question from computational neuroscience: What is the computational benefit of the astounding diversity of neuron types in the neocortex that have emerged during brain evolution?

\clearpage

\section*{Methods}
\label{sec:methods}
\subsection*{Neuron model with after-hyperpolarizing (AHP) currents}
\label{sec:methods-LIF-desc}

The dynamical behavior of an LIF-neuron and the AHP-neuron that has after-hyperpolarizing (AHP) currents (indexed by $j$), as implemented in Loihi, is given by Eq.~\ref{eqn:methods-AHP-defintion}--\ref{eqn:voltage-reset}.
Here we show the dynamic interaction between incoming spikes at time $t$, the resultant input postsynaptic current (PSC) $\ipscj[t]$, the internal AHP-current $\iAHPj[t]$, the membrane voltage $V_j[t]$, and the output spikes $z_j[t + 1]$. The equations are explained subsequently
\mathcenter
\begin{align}
    \ipscj[t+\deltat] &= \alpha_I\; \ipscj[t] + \sum_{i} w_{ij}\, z_i[t - d_{ij}] \,, \label{eqn:methods-PSC-definition} \\
    \iAHPj[t+\deltat] &= \alpha_{AHP}\; \iAHPj[t] - \beta\; z_j[t] \,, \label{eqn:methods-AHP-defintion} \\
    V_j[t+\deltat] &= \left\{\begin{array}{ll}
        \alpha_V V_j[t] + \frac{1}{g_V}(\ipscj[t+\deltat] + \iAHPj[t+\deltat]) & \text{if neuron is not refractory} \\
        0 & \text{otherwise}
    \end{array}\right. \,, \label{eqn:methods-LIF-membrane-voltage}\\
    z_j[t+\deltat] &= \left\{\begin{array}{rl}
        1 & \text{if }V_j[t+\deltat] > b_0 \\
        0 & \text{otherwise}
    \end{array}\right. \,, \label{eqn:methods-spike-threshold-crossing} \\[13pt]
    V_j[t+\deltat] &\rightarrow 0 \text{ if } z_j[t+\deltat] = 1 \,.\label{eqn:voltage-reset}
\end{align}
Eq.~\ref{eqn:methods-PSC-definition}--\ref{eqn:methods-LIF-membrane-voltage} implement temporal convolution with an exponentially decaying kernel.
Here $\alpha_I = e^{-\frac{1}{\tauI}}$, $\alpha_{AHP} = e^{-\frac{1}{\tauadap}}$, $\alpha_V = e^{-\frac{1}{\tauV}}$, where $\tauI$, $\tauadap$, and $\tauV$ are the decay constants of the corresponding exponentials.
$\beta$ is the update to the AHP-current in response to an output spike. Since the state transition is computed in Loihi, we associate a single compute step with 1ms biological time, correspondingly $\deltat=1\ms$ and $g_V=1$.

Eq.~\ref{eqn:methods-PSC-definition} defines the PSC as a function of input spikes arriving through incoming synapses of weights $w_{ij}$ and delays of $d_{ij}$ steps.
The LIF-neuron that is without AHP-currents corresponds to the case of $\beta = 0$, where the neuron performs a leaky-integration of $\ipscj[t]$ to get the membrane voltage $V[t]$.
When this voltage exceeds a threshold, it is reset to zeros and an output spike is generated.
In this case, the memory of the neuron is limited by the voltage and PSC decay time constants $\tauV$ and $\tauI$ respectively, which are typically around $20\ms$.
This means that even when connected in a recurrent fashion, the memory capacity for the network is typically at most a $100\ms$.

Eq.~\ref{eqn:methods-AHP-defintion} defines the AHP-current.
With $\beta > 0$, each output spike i.e. $z_j[t] > 0$ will cause $\iAHPj[t]$ to become more negative by a value of $\beta$.
When leaky-integrated into the membrane voltage $V[t]$ (Eq.~\ref{eqn:methods-LIF-membrane-voltage}), this increased negative value of $\iAHP[t]$ lowers the rate of subsequent spikes, leading to \emph{spike frequency adaptation}.
The decay time constant of the AHP-current $\tauadap$ is much longer than $\tauV, \tauI$, typically $>100\ms$.
The slow decay means that this inhibition persists over a much longer duration, thus functioning as a longer-term memory cell.
This longer lasting memory proves invaluable to solve the complex tasks demonstrated in this work.

This model can be very efficiently implemented in Loihi making use of its support for multi-compartment neuron models.
The AHP-current is calculated in the voltage register of a separate compartment.
Loihi then allows this voltage to be integrated into the voltage of the main compartment as an additional current.
This usage of multiple compartments to implement slow-changing dynamics differentiates our model from all previous neuromorphic implementations of memory.

\subsection*{Details of spiking neural network training}
\label{sec:methods-ALIF-training}

    In this section we describe important details pertaining to the training of networks of LIF-neurons and AHP-neurons. In all equations below, we drop the neuron index $j$ for brevity.

    \subsubsection*{The scaled voltage}
        For subsequent details regarding SNN training, we find it useful to define a normalized version of the membrane voltage i.e. a \emph{scaled voltage $v_s$}.

        We first notice that the membrane voltage $V[t]$ is a sum of two voltage components $\vPSC[t]$ and $\vAHP[t]$ which are a result of leaky-integrating $\ipsc$ and $\iAHP$ respectively. Correspondingly, we can rewrite Eq.~\ref{eqn:methods-LIF-membrane-voltage},\ref{eqn:voltage-reset} describing the voltage evolution as follows:
        \begin{align}
            \vPSC[t+\deltat] &= \left\{\begin{array}{ll}
                \alpha_V \, \vPSC[t] + \frac{1}{g_V}\ipsc[t+\deltat] & \text{if neuron is not refractory} \\
                0 & \text{otherwise}
            \end{array}\right. \,, \label{eqn:methods-vPSC-def}\\
            \vAHP[t+\deltat] &= \left\{\begin{array}{ll}
                \alpha_V \, \vAHP[t] + \frac{1}{g_V}\iAHP[t+\deltat] & \text{if neuron is not refractory} \\
                0 & \text{otherwise}
            \end{array}\right. \,, \label{eqn:methods-vAHP-def}\\
            V[t+\deltat] &= \vPSC[t+\deltat] + \vAHP[t+\deltat] \,.
        \end{align}
        \begin{equation}
            \begin{array}{rl}
            \vPSCj[t+\deltat] \rightarrow 0 \\
            \vAHPj[t+\deltat] \rightarrow 0
            \end{array} \text{ if } z_j[t+\deltat] = 1  \label{eqn:voltage-reset-split}    
        \end{equation}

        The scaled voltage $v_s[t]$ is defined below:
        \begin{equation}
            v_s[t] = \frac{V[t] - b_0}{b_0 - \vAHP[t]} \label{eqn:scaled-voltage-def}
        \end{equation}

        $v_s[t]$ takes the value of 0 when $V[t] = b_0$ and a value of $-1$ when $V[t] = \vAHP[t]$. This is motivated by the fact that $\vAHP[t]$ is the value that $V[t]$ would take if there was no input PSC.

    \subsubsection*{The surrogate gradient}
        The generation of spikes from the membrane voltage (Eq.~\ref{eqn:methods-spike-threshold-crossing}), involves the use of a step function centered at the neuron threshold.
        This function is non-differentiable at the neuron threshold and provides a non-informative gradient of zero at all other points.
        Thus in order to use gradient back-propagation to train networks of spiking neurons, we consider a surrogate gradient for the step function similar to methods used in previous works
        \cite{bellec2018,
        zenke2021remarkable,
        esser2016convolutional,
        shreshtha2018slayer,
        neftci2019surrogate,
        zenke2018superspike,
        zhu2021efficient}.

        We rewrite the thresholding equation Eq.~\ref{eqn:methods-spike-threshold-crossing} in terms of the scaled voltage $v_s[t]$:
        \begin{equation}
            z[t+\deltat] = h\robrace{v_s} \equiv h\robrace{\frac{V[t] - b_0}{b_0 - \vAHP[t]}} \,,
            \label{eqn:spike-step-function}
        \end{equation}
        where $h$ is the unit step function.

        We then use the following piece-wise linear surrogate gradient function to serve as a pseudo-derivative of the step function $h(\cdot)$.
        \begin{equation}
            \frac{dh}{d v_s} \triangleq \left\{\begin{array}{rl}
                \gamma\robrace{1 + \frac{v_s}{v_-}} & \text{if } -v_- \le v_s < 0 \\
                \gamma\robrace{1 - \frac{v_s}{v_+}} & \text{if } 0 \le v_s \le v_+ \\
                0 & \text{otherwise}
            \end{array}\right. \,,
            \label{eqn:pseudo-derivative-def}
        \end{equation}
        where $v_-$ and $v_+$ define the support of the surrogate gradient and $\gamma$ is a dampening factor that affects the magnitude of the derivative.

        Thus $h'(v_s)$ peaks at a value of $\gamma$ for $v_s = 0$ and linearly decays to zero at the values of $-v_-$ and $v_+$.

    \subsubsection*{Impact of AHP-neurons on gradient propagation}
        In this section we provide the theory that explains how gradients are protected and propagated further back in time with the presence of AHP-neurons.
        In this case, we consider the gradient of a loss $L$ that depends only on the output spikes $z[t]$.

        For a standard LIF-neuron, the following three equations govern gradient back-propagation in time.
        For a neuron indexed $i$, the gradient $L$ with respect to the state variables propagates as follows.
        \begin{align}
            \dL{V_i[t]} &= \alpha_V \dL{V_i[t+\Delta t]} + \partd{z_i[t]}{V_i[t]} \dL{z_i[t]} = \boxed{\alpha_V \dL{V_i[t+\Delta t]}} + \frac{h'(v_{s,i}[t])}{b_0} \dL{z_i[t]} \label{eqn:grad-prop-V} \,,\\
            \dL{\ipsci[t]} &= \alpha_I \dL{\ipsci[t+\Delta t]} + \partd{V_i[t]}{\ipsci[t]} \dL{V_i[t]} = \boxed{\alpha_I \dL{\ipsci[t+\Delta t]}} + \frac{1}{g_V} \dL{V_i[t]} \label{eqn:grad-prop-IPSC} \,, \\
            \dL{z_i[t]} &= \partd{L}{z_i[t]} + \boxed{\sum_{j} w_{ij}\, \dL{\ipscj[t + d_{ij} + \Delta t]} \label{eqn:grad-prop-z}} \,.
        \end{align}

        In Eq.~\ref{eqn:grad-prop-V}, we see the gradient decay proportional to $\alpha_V$ through the membrane voltage.
        In Eq.~\ref{eqn:grad-prop-IPSC}, we see a decay proportional to $\alpha_I$ through the PSC. Both decay factors correspond to small decay time constants (5-20ms) and thus don't preserve gradient information too far back.
        Additionally the gradient propagates backwards from spikes to voltages to the PSC and then through the recurrent weights, by applying Eq.~\ref{eqn:grad-prop-V}--\ref{eqn:grad-prop-z} in sequence.
        The decay of the gradient through this channel is proportional to the pseudo-derivative $h'(v_{s,i}[t])$ and thus the dampening factor $\gamma$.
        $\gamma$ is typically kept low because gradients that propagate via the pseudo-derivative are approximations and become less informative the further they propagate.
        Additionally this helps keep the gradient propagation stable.

        All of the above mean that the informative gradients don't propagate too far back in time. However, with the addition of the slow-decaying AHP dynamics, we have the following:
        \begin{align}
            \dL{\iAHPi[t]} &= \alpha_{\text{AHP}} \dL{\iAHPi[t+\Delta t]} + \partd{V_i[t]}{\iAHPi[t]} \dL{V_i[t]} = \boxed{\alpha_{\text{AHP}} \dL{\iAHPi[t+\Delta t]}} + \frac{1}{g_V} \dL{V_i[t]} \,, \label{eqn:grad-prop-IAHP}\\
            \dL{z_i[t]} &= \boxed{-\beta \dL{\iAHPi} + \partd{L}{z_i[t]} \sum_{j} w_{ij}\, \dL{\ipscj[t + d_{ij} + \Delta t]} \,. \label{eqn:grad-prop-z-with-AHP}}
        \end{align}

        Eq.~\ref{eqn:grad-prop-IAHP} shows a means for gradients to propagate through the AHP-currents with a very slow decay $\alpha_{\text{AHP}}$, which allows the propagation of an informative gradient signal far back in time. This gradient signal is then propagated through the spikes (Eq.~\ref{eqn:grad-prop-z-with-AHP}) and consequently through the voltage and current and the weights.

        Fig.~\ref{fig:sMNIST} D demonstrates the impact that the AHP-current has over the propagation of $\dL{z[t]}$ over time.
        In the top plot, the slowly decaying gradient is visible in the $\left|\dL{z[t]}\right|$ of the AHP-neurons.
        This gradient then gives an informative gradient signal to the rest of network.

    \subsubsection*{Spike rate regularization}
        For each neuron $k$, we calculate the mean rate $\bar{\rho_k}$ across all batches. We then add the following regularization loss
        \begin{equation}
            L\rho = \lambda_\rho(\sum_j(\bar{\rho_k} - \rho_{target})^2)^2 \,,
        \end{equation}
        where $\rho_{target}$ is a target rate and $\lambda_\rho$ is the parameter that controls the strength of the regularization.
        This loss encourages the mean spike rate of each neuron across a random batch to be as close as possible to the target rate $\rho_{target}$.
        This ensures that the network activity does not die out and that the spike rate stays sparse owing to the low value of $\rho_{target}$.
        The outermost square is in order to dynamically reduce the regularization strength as the loss becomes smaller.

        When training the Spiking RelNet, we use a more aggressive spike rate regularization to limit the total spike rate across all the instances of the relational function $g_\theta$.
        This is described below in the section on the training of the Spiking RelNet.

\subsection*{Details for calculation of pointwise separation (PSPR)}

We considered two networks with randomly drawn weights, one with and one without AHP-neurons. In each case we measured the extent by which images from two classes (handwritten 6 and 8) are separated by the spiking activity of the network after the input sequences have been fed it. More precisely, we considered the 240-dimensional vector consisting of the low-pass filtered spikes (filter time constant: $20\ms$) from all network neurons at time $T=840\ms$. At that time point not only have the input sequences ended, but also spikes from a separate input neuron have been provided from time $784\ms$ to $840\ms$. These serve as a "prompt" that transforms values of hidden variables of the network states, the current values of AHP-currents, into observable spiking activity. We normalized these vectors to 1, and then measured the Euclidean distance between any two such vectors that resulted from handwritten samples of the digits 6 and 8. 
We used 100 MNIST~\cite{lecun2010mnist} images from each of the two classes.
Each image was presented as described in the section on Input encoding for the sMNIST task. The histogram of the resulting pairwise distances is plotted in Fig.~\ref{fig:sMNIST} C.

To mirror our sMNIST solution we used a network of 240 LIF-neurons and compared it with the AHP-SNN with 140 LIF-neurons and 100 AHP-neurons. 

In order to test how well the vectors that result from the two input digits 6 and 8 could be separated by a linear classifier, we trained a standard max-margin linear support vector classifier. We used 500 images per class.

\subsection*{Details for the application to sMNIST}

\subsubsection*{Input encoding}
The gray values of the pixels from an MNIST~\cite{lecun2010mnist} image were encoded in spikes using a threshold crossing encoding (see Fig.~\ref{fig:sMNIST} A. The number of thresholds was chosen to be 40, which is half of the 80 input neurons. So we had 2 neurons for each threshold, one spiking when the gray value between the former and the new pixel increased and the other neuron spiking when the gray value decreased by transitioning to the new pixel. The thresholds were linearly spaced between 0 and 255. The pseudo code for the input encoding can be seen in the Supplement.
After the presentation of all 784 pixels, which take $1\ms$ each, the $80^{th}$ input neuron becomes active for an additional $56\ms$. Thus, the presentation of one sample takes $840\ms$. This last input neuron which generates a spike at every time step, for 56 steps after the image presentation, indicates the end of a sample. The classification happens at the last time step i.e. time step 840 of a sample. Each of the 10 output neurons denotes a digit and the neuron with the highest membrane potential on the last time step defines the predicted class. The network was implemented on the Intel Loihi chip using NxNet API from the NxSDK v0.95.\\

\subsubsection*{Network structure}
An AHP-SNN was used consisting of 240 neurons, 180 excitatory and 60 inhibitory. A random subset of 100 of the excitatory spiking neurons were AHP-neurons. Additionally, 80 input neurons were used to perform an input spike encoding of the images, and 10 output neurons were used corresponding to the 10 classes of the MNIST dataset. The hyper-parameters which were used to train the network for Loihi were $\beta = 96$, baseline threshold $b_0 = 127$, $\tau_V = 20\ms$, $\tauadap = 700\ms$ as well as a refractory period and delay of $1\ms$.

\subsubsection*{Control models}
Given here is an overview of the network structures from Bellec et al.~\cite{bellec2018} to which the AHP-SNN was compared in Fig. \ref{fig:sMNIST} E for solving the sMNIST task. The SNN without AHP-neurons consists of 220 LIF-neurons with 100\% connectivity, the LSNN consists of 120 LIF-neurons and 100 neurons with adaptive thresholds with 12\% connectivity, the recurrent ANN without LSTM units consists of 128 hidden units with hyperbolic tangent activation function with 100\% connectivity and the ANN with LSTM units consists of 128 LSTM units with 100\% connectivity.

\subsubsection*{Training details}
The AHP-SNN for solving sMNIST was trained for 465 epochs on the MNIST\cite{lecun2010mnist} training set during two stages similarly as in Akopyan et al.~\cite{truenorth2015}. The forward pass uses a software model of Loihi neurons including the limited precision on neural states, e.g., 8 bits for weights. For the backward pass the full precision weight matrix was kept to calculate the weight updates, which were quantized afterwards for the forward pass. During training, the overall connectivity of the network was kept at 20\%, including the input and output connectivity. So only 20\% of the possible synapses between the neurons were active. This was achieved by using the rewiring technique DEEP-R~\cite{bellec2018deep} during training, which is based on random synaptic sampling. Constraining the connectivity to 20\% of the available connections was needed to stay within the limited axonal and synaptic resources of a Loihi neuro-core. The resulting sparsely connected AHP-SNN had an average firing rate of 29Hz per neuron on Loihi classifying the test set of MNIST~\cite{lecun2010mnist}.

\subsection*{Spiking RelNet Network Architecture}

    \label{sub:network_structure}

    In this section, we describe in detail the structure of the Spiking RelNet as applied to the bAbI tasks.

    Building on the general architecture proposed in \cite{santoro2017}, the Spiking RelNet takes as input $K$ objects $o_i(t)$, and a question object $q(t)$ and implements the following function to compute its output.
    \begin{equation}
        RN\robrace{\sqbrace{o_1(t), o_2(t), \ldots, o_K(t)}, q(t)} = f_\phi\robrace{f_{agg}\robrace{\sum_{1 \le i \le j \le K}{g_\theta\robrace{o_i(t), o_j(t), q(t)}}}} \,.
        \label{eqn:methods-spiking-relnet-definition}
    \end{equation}

    Fig.~\ref{fig:relational_network} A shows the basic outline of this network. 
    When applied to the bAbI task, the sentences of the story and the question are embedded into the spike sequences $o_i(t)$ and $q(t)$ respectively by means of AHP-SNNs, which are recurrent networks consisting of LIF-neurons and AHP-neurons.
    To provide the input to the AHP-SNN, we assigned an input neuron corresponding to each distinct word used in the bAbI dataset.
    The words in a sentence/question were then presented in sequence, with each word being presented for a duration of $\Tword = 10\ms$ during which only the corresponding input neuron fired continuously.
    We then took the spike activity of the AHP-SNN over the last $\tinp = 14\ms$, and padded it to a length of $\tsim = 37\ms$ to form the embedding spike sequences $o_i(t)$ and $q(t)$ (see Fig.~\ref{fig:relational_network} B).

    The function $g_\theta$ is the relational function.
    It receives as input a triplet of spike sequences $(o_i(t), o_j(t), q(t))$ corresponding to a pair of sentences and questions, and produces a spike sequence output
    It is implemented as a four layer feed-forward spiking neural network with LIF-neurons.
    We have an instance of $g_\theta$ for each pair of sentences $i, j :: i \le j$, so that the ordering of the sentences in the stories is made available to the network.

    The function $f_{agg}$ is an element-wise function. 
    This is implemented by means of a LIF-neuron layer to which each instance of $g_\theta$ is connected one-to-one, where the set of input weights from an instance of $g_\theta$ to this layer is shared across all instances.
    This is an addition to the architecture proposed in Santoro et al. \cite{santoro2017} and plays an essential role in enabling the implementation of Spiking RelNets onto neuromorphic hardware.

    The function $f_\phi$ is the readout function.
    It is implemented as a 3-Layered feed-forward LIF-neuron network followed by a linear readout (see section below) and a softmax layer.
    $f_\phi$ outputs, for each unique word present in the bAbI dataset, the probability of that word being the answer to the question.
    This probability is used to compute a cross-entropy loss that is used to train the network using gradient back-propagation.
    For more detailed parameters pertaining to the layers, see Supplement.

\subsection*{Optimizing performance and energy efficiency of Spiking RelNet}
    The network outline above shows that the Spiking RelNet uses both recurrent networks for embedding (part B of Fig.~\ref{fig:relational_network} A) as well as feed-forward networks for the relational function and readout (Parts C, D, E of Fig.~\ref{fig:relational_network} A) to perform the calculation.
    When scaling up to tasks with a larger number $K$ of sentences, the fraction of these feed-forward components of the Spiking RelNet increases (See Fig.~\ref{fig:regularization-stats} B). The reason is that the number of sentences to be embedded, and hence AHP-SNN Networks, scales linearly with $K$, whereas the number of feed-forward modules that compute the function $g_\theta$ increase quadratically with $K$, since we have an instance of $g_\theta$ for each pair of sentences. Consequently, the numerous instances of the relational function ($g_\theta$) occupy the majority of the hardware resources (see Fig.~\ref{fig:board} C). This increasing fraction of feed-forward network modules is problematic from the perspective of energy efficiency, since prior emulations of feed-forward networks in spike-based hardware demonstrated that their advantage regarding energy-consumption gets lost for larger networks when high classification accuracy is required~\cite{davies2021}. The reason is that these prior implementations had to use spike rate coding instead of event-based processing in order to achieve high classification accuracy. However, rate coding uses many spikes per neuron, thereby moving the network out of an energy-efficient working regime, and also increases the computation time of the network, i.e., reduces its throughput.

    Additionally, the Spiking RelNet requires more compute time (time steps) compared to the non-spiking RelNet, making the loss computation and gradient back-propagation through time many times more expensive in the spiking case.
    Simply training the network end-to-end with the cross-entropy loss requires an impractically long time for the network to converge, as well as leading to pathological spike rates and low performance. The solutions to these issues are:

    \subsubsection*{Pre-training the AHP-SNN}

        In order to speed up the convergence of the training algorithm, we first trained a non-spiking relational network to solve the bAbI tasks, where LSTMs were used to embed the questions and words.
        We then trained the AHP-SNN to reproduce the outputs of the LSTMs for the various input sentences in the dataset. 
        The weights of these pre-trained AHP-SNNs were fixed, and they were used to perform the embedding while we trained the relational function $g_\theta$ and readout function $f_\phi$.

        The AHP-SNN for sentences were pre-trained for 3500 Epochs with 5218 data points per epoch. For the questions 38000 Epochs with 478 points per epoch were used. 
        Using the pre-trained LSNN, training for the Spiking RelNet in contrast takes only 80 Epochs, with 17000 data points per epoch, demonstrating the faster convergence achieved due to pre-training.
        Despite the larger number of epochs, pre-training the embedding was cheaper (15 hours using a single GPU) due to the much larger size of the Spiking RelNet (which took over a week on 2 GPUs for 80 Epochs).
        The resultant fewer training epochs needed compared to the end-to-end trained non-spiking RelNet makes the training feasible for a Spiking RelNet.

    \subsubsection*{Compressing the network in time}
        Another aspect of optimization concerns the number of time steps used, which represents the compute time, which affects not only the energy consumed and delay on Loihi, but also the training speed. We found that using spiking neurons without refractory period and membrane time constants $\tauadap$ of just 7 ms significantly reduced the required number of time steps, while causing only a mild decrease in accuracy. In the Supplement we show that a network whose dynamics are stretched over more time steps can solve all 17 tasks at the cost of higher energy consumption and latency.

    \subsubsection*{Aggressive spike-rate regularization and voltage regularization}
        The energy inefficiency of DNNs due to higher spike rates, which is a common experience with CNNs, can be overcome in the case of the Spiking RelNet.
        An important underlying difference to CNNs is that even for large problem instances, i.e., stories with many sentences,
        the number of relations that are relevant for answering a question tends to increase only linearly with the length of the story.

        We used a more aggressive regularization for the spike rates in the instances of the relational function $g_\theta$, where the regularization forces the total spike rate summed over all instances of $g_\theta$ towards a low target rate.
        This forced the network to allocate spiking activity only for those instances of $g_\theta$ that corresponded to relevant pairs of sentences.
        The resultant low spike rate seen in Fig.~\ref{fig:regularization-stats} B resulted in a very power and delay efficient implementation of feed-forward spiking networks onto neuromorphic hardware.
        See Supplement for the precise formulation and more details.

    \subsubsection*{Voltage regularization}
        The strong spike rate regularization described above has a tendency to push the synaptic weights low enough that the membrane voltages become very negative. This leads to a large number of time steps where the voltage values fall outside the support of the surrogate gradient and thus no gradient information can be propagated through them, which impedes gradient back-propagation.
        Thus we were motivated to add a loss that penalizes voltages that fall significantly outside the support of the surrogate gradient function defined in Eq.~\ref{eqn:pseudo-derivative-def}.
        Since the surrogate gradient is defined in terms of the scaled voltage $v_s$ (Eq.~\ref{eqn:scaled-voltage-def}), we define the voltage regularization loss in terms of it as well.

        For each neuron $j$ and time step $n$, we calculate the loss component
        \begin{equation}
            L_v^{(j, n)} = (\mbox{relu}(v_{s, j}[n] - 0.4))^2 + (\mbox{relu}(-v_{s,j}[n] - 2.0))^2 \,.
        \end{equation}
        The total voltage regularization loss is given by
        \begin{equation}
            L_v = \lambda_v\robrace{\mean_{i, n} {L_v^{(i, n)}}}^2 \,.
        \end{equation}
        
        This loss was chosen with the support of the surrogate gradient $v_s \in [v_-, v_+] = [-1.0, 0.5]$ in mind. Hence the above penalizes all neurons at all time instants where the scaled voltage $v_s$ goes outside the range $[-2.0, 0.4]$. This prevented the network from using voltages that are excessively negative and increases the proportion of voltage values that lie within the support of the surrogate-gradient.
        Moreover, limiting the range of the voltage values was also crucial in order to be able to fit the voltage values onto the range offered by the fixed precision registers on Loihi.

    \subsubsection*{The linear readout}

        The design of the linear readout was crucial to the performance of the relational network.
        The linear readout consists of a network of specialized readout neurons, with one neuron for each word in the database of words used in the bAbI task.

        The readout neuron is a variant of the standard LIF-neuron, where $\tauI = \taureadout = 7.0\ms$ and $\tauV = \infty$, and the threshold $b_0 = \infty$.
        This corresponds to a neuron which does not spike, but where the PSC decays with the readout time constant $\taureadout$ and the neuron performs (non-leaky) integration of the PSC to calculate the membrane potential.
        However, we chose to enable the integration of the PSC into the voltage only $\Treadout=10\ms$ prior to the final step.
        The value of the membrane voltage at the final step is scaled by a fixed scalar and forms the input to the softmax (see Fig.~\ref{fig:relational_network} E).
        This design incentivizes the spike activity of the final layers to occur in a confined time window close to the final time step,
        while allowing the precise timing of the spike to influence the final output, leading to a high information capacity in a short time window.

\subsection*{Placement of the Spiking RelNet onto Loihi}
    The Loihi Nahuku board consists of 32 interconnected Loihi chips, each of which contains 128 neuro-cores. The neuro-core is the fundamental computational unit that computes the dynamics of the LIF-neurons and AHP-neurons. Loihi allows one to connect any neuro-core on any chip to any other neuro-core on any other chip, thus enabling large networks to be placed on the board. However due to hardware limitations, the number of connections and the connectivity is constrained as described below. Additionally, transporting a large number of spikes across different chips incurs significant latency. We discuss here the strategies to place the Spiking RelNet within these constraints.

    The AHP-SNN network that solves the sMNIST task contains 240 neurons connected with 20\% of the recurrent connections enabled. This network is small enough to fit in a single chip and occupy only one neuro-core.
    The Spiking RelNet is a much larger network.
    Considering a maximum of $M=20$ sentences in a story, the Spiking RelNet has $M$ instances of the AHP-SNNs that embed sentences, plus one for the questions.
    Additionally, there exists an instance of the relational function $g_\theta$ for each pair of sentences $o_i(t), o_j(t) :: i\le j$, making a total of $\frac{M(M+1)}{2} = 210$ instances.
    Each of these instances is implemented as a separate network on Loihi, leading to a total network size of $238,604$ neurons.
    The placement of this network needs to take into consideration many constraints regarding connectivity, memory, and the latency of spike transport.
    The associated challenges and solutions are outlined in this section.

    \subsubsection*{Synaptic memory limit}
        Each neuro-core has a limited amount of SRAM memory which can be used to store synaptic parameters.
        This limits the number of incoming synapses to a particular neuro-core.
        The precise number is dependent on synaptic parameters and we have found an empirical limit of around 40000 synapses per neuro-core.
        Except the aggregation layer, all layers in the network have dense input and recurrent synaptic connections.
        Thus each layer needs to be placed over multiple neuro-cores in order to store the input and recurrent connections.

    \subsubsection*{Fanout limits -- AHP-SNN relay layer}
        The total number of neuro-cores to which the neurons of a neuro-core connect to is limited to 2048, and 4096 for intra-chip connections.
        This plays a role when connecting the AHP-SNNs to the large number of instances of the relational function $g_\theta$.

        Thus, one can split the neurons across multiple neuro-cores to reduce the number of output connections per neuro-core.
        However splitting a recurrent AHP-SNN network across too many neuro-cores increases latency.
        Instead we use relay layers.
        A \emph{relay layer}, as the name suggests, simply reproduces the spiking activity of the layer that forms its input.
        Each AHP-SNN is thus connected to multiple relays which then each fanout to a smaller number of instances of $g_\theta$.

    \subsubsection*{Limits pertaining to fanin -- The aggregation layer}
        For any neuro-core $C$, Loihi limits the number of neurons that can be connected to that neuro-core to $4096$.
        Unlike the two constraints above, this constraint on the fanin to a neuro-core introduces a fundamental restriction to the network architectures that can be implemented on Loihi.

        The layer that receives the output from the instances of $g_\theta$ receives input from $\frac{M(M+1)}{2} = 210$ instances.
        For this layer to not violate the fanin constraint, the connection from the output of $g_\theta$ to this layer must be sparse.
        Thus, we introduce an \emph{aggregation layer} to which each instance of $g_\theta$ is connected in a sparse one-to-one manner, with shared weights across instances.
        The sparse connection enables the aggregation layer to be implemented within the fanin constraints

    For a more detailed treatment of the constraints, as well as the number of neuro-cores required to place each layer, see Supplement.

    \subsubsection*{Optimizing network placement to minimize congestion in cross chip spike transport}

        Placing the AHP-SNN, relay, and the relational networks taking into consideration only the connectivity constraints, we noticed significant delays that occured due to transporting spikes from the AHP-SNN and relay networks to the instances of $g_\theta$.
        This is owing to the large number of spikes that need to be transferred across different Loihi chips.
        Thus in addition, we needed to optimize the placement of the instances of $g_\theta$, and the relay networks in a manner that minimizes cross-chip spike transport.
        We broke down this general objective into the following constraints.
        \begin{itemize}
            \item All relay networks must be connected only to relational function instances that are placed on the same Loihi chip.
                  We thus choose to place the initial layer of the relational function instances in the same chip as the relay networks that give them their input.

            \item We aim to minimize the number of relay networks required.
                  Each chip has a limit of $128$ neuro-cores and thus a limit on the number of $g_\theta$ instances that can be placed.
                  This means that for each chip, we must choose the set of $g_\theta$ instances in such a manner that the number of distinct sentences needed as input is minimized.

            \item For each $g_\theta$ instance, all layers after the first one are to be placed on the same chip.
        \end{itemize}
        The layout that we arrived at with the above principles is described in the Supplement.
        The resultant improvement in delay and the corresponding energy delay product is shown in Fig.~\ref{fig:board} E.

\section*{Data availability}
The MNIST dataset~\cite{lecun2010mnist} is freely available at \url{http://yann.lecun.com/exdb/mnist/}.
The bAbI dataset~\cite{weston2015} is freely available at \url{https://research.fb.com/downloads/babi/}.

\section*{Code availability}
The Loihi source code is freely available from Github (\url{https://github.com/intel-nrc-ecosystem/models/tree/master/nxsdk_modules_ncl/lsnn/}).

\bibliography{bib/bib}

\section*{Acknowledgements}

This research/project was supported by the Human Brain Project (Grant Agreement number 785907 and 945539)
of the European Union and a grant from Intel. Special thanks go to Guillaume Bellec and Darjan Salaj for their insightful comments and ideas when carrying out this work.

\section*{Author contributions statement}

A.R., P.P. and W.M. contributed to the design and planning of the experiments. A.R. and P.P. carried out the experiments. A.R., P.P., A.W. and W.M. participated in the analysis of the experimental data. A.R., P.P., A.W. and W.M. wrote the manuscript.

\section*{Competing interests}

The authors declare competing interests as follows. P.P. and A.W. are currently employed by Intel Labs, developers of
the Loihi neuromorphic system. W.M. and A.R. are members of the Intel Neuromorphic Research
Community and W.M. has received research funding from Intel for related work.

\section*{Additional information}
Supplementary information is available.

\end{document}


\maketitle

\tableofcontents

\listoftables

\listoffigures

\section{Neuron models}

    \subsection{Neuron Parameters}
        The parameters pertaining to the neurons used in the AHP-SNNs are listed in Table~\ref{tab:neuron-params}.
\begin{table}[!h]
\edef\oldaboverulesep{\aboverulesep}
\edef\oldbelowrulesep{\belowrulesep}
\renewcommand{\aboverulesep}{0.0ex}
\renewcommand{\belowrulesep}{0.0ex}
\centering
\begin{tabular}{>{\centering}p{3ex}r|c|cc}
\toprule
 &  &  & \multicolumn{2}{c}{Spiking RelNet}\\
\cmidrule(lr){4-5}
\multicolumn{2}{l|}{Parameter} & sMNIST & Original & Stretched in time\tabularnewline
\hline 
\multicolumn{2}{l|}{\textbf{Neuron Parameters:}}             &       &        &        \tabularnewline
 & PSC decay $\tauI$ (steps)                                 & 0.0   & 7.0    & 20.0   \tabularnewline
 & Voltage decay $\tauV$ (steps)                             & 20.0  & 7.0    & 20.0   \tabularnewline
 & AHP-current decay $\tauadap$ (steps)                      & 700.0 & 40.3   & 120.0  \tabularnewline
 & AHP-current decrement $\beta$ / $V_{\text{thr}}$          & 0.756 & 0.176  & 0.062  \tabularnewline
 & Refractory period $\tref$                                 & 0   & 0      & 2      \tabularnewline[2ex]

\multicolumn{2}{l|}{\textbf{Surrogate gradient Parameters:}} &  &                         &                        \tabularnewline
 & Dampening factor $\gamma$                                 & 0.3 & 0.0                     & 0.5                  \tabularnewline
 & Scaled voltage support $\left[v_{-},v_{+}\right]$         & $\left[-1.0,1.0\right]$  & $\left[-1.0,0.5\right]$ & $\left[-1.0,0.5\right]$ \tabularnewline
\bottomrule
\end{tabular}
\renewcommand{\aboverulesep}{\oldaboverulesep}
\renewcommand{\belowrulesep}{\oldbelowrulesep}
\caption[LIF-Neuron and AHP-neuron parameters]{\textbf{LIF-neuron and AHP-neuron parameters:}~ Here we detail the parameters for the spiking neurons used in the different examples.
In each task, the parameters for the AHP-neurons are identical to those of the LIF-neurons, with the exception of the additional parameters $\beta$ and $\tauadap$. The examples for which the parameters are shown are: the sMNIST task solved by a single AHP-SNN network, the "original" version of the Spiking RelNet on the bAbI task that is benchmarked in the main text, and a Spiking RelNet where all parameters are adjusted to effectively slow the simulation down, and increase temporal resolution by a factor of 3 (for which the results are shown in Table~\ref{tab:bAbI-results})}
\label{tab:neuron-params}
\end{table}

\section{Energy and Time benchmarking}

    In order to evaluate the energy efficiency of our spiking networks implemented on the neuromorphic chip Loihi from Intel~\cite{davies2018} we measured the energy and latency of a task and compared it with the artificial neuronal network implementation on conventional hardware, i.e., CPUs and GPUs.

    For the network executed on a Loihi system the performance is measured by reading out sensors of the hardware. Regarding the energy measurement, Loihi chips of a system are powered by on-board voltage regulators that support power telemetry over an I²C interface. These voltage regulators are used to collect power usage information. In particular, the SDK of Loihi allows to split the contributions of static and dynamic power consumption as well as estimate the contribution of neuro-cores and on-chip synchronous x86 cores to the overall power consumption.

    For Energy measurement on CPU the Intel Power Gadget 3.5, a software based power estimation tool, was used. For GPUs we used the nvidia-smi tool to measure the power, which is also a software based estimation tool. The nvidia-smi tool does not give a detailed breakdown of where power is consumed, but rather report the power draw of the whole board. Therefore we measured a baseline idle power draw, which we considered for the static energy. Afterwards we measure the power consumption during the workload, which denotes to the total energy and then we calculated the dynamic energy by subtracting the static energy from the total energy.

    For both Loihi and CPU/GPU the measurements were performed for a workload running long enough to get in a steady state for power draw. Therefore, the batch size 50 and 100 examples on the GPU required us to run the test set several times to achieve a steady state. The execution time for networks executed on Loihi and networks running on CPU or GPU was measured on python level using the timeit module and can be considered a wall-clock time. This wall-clock time is then divided by the number of samples used for inference to calculate the latency. The execution time was measured independent of the power measurements.

\section{Details for AHP-SNN on the sMNIST task}

    \subsection{Input encoding for sMNIST}

        In Listing~\ref{lst:smnist-input-code} the pseudo code for the input encoding used in the sMNIST task is shown. We assume that the current pixel value and next pixel value of the input image are presented, the number of thresholds were chosen to be half of the input neurons and thresholds are linearly spaced between 0 and 255 (number of threshold times).

        \begin{lstlisting}[
            caption={\textbf{Input encoding sMNIST.}~Pseudo code for the input spike encoding of MNIST images used in the sMNIST task.},
            label={lst:smnist-input-code},
            language=listinglang,
            captionpos=b,
            basicstyle=\fontfamily{pcr}\selectfont\footnotesize,
            keywordstyle=\bfseries,
            commentstyle=\itshape,
            rulecolor=\color{black},
            frame=single,
            float=h!]
while threshold_counter <= num_thresholds:

    thr = thresholds[threshold_counter]

    # transition from a lower pixel value to a higher pixel value
    if current_pixel_value <= thr and next_pixel_value >= thr:
        input neuron with the id 2*threshold_counter spikes
    
    # transition from a higher pixel value to a lower pixel value
    if current_pixel_value >= thr and next_pixel_value <= thr:
        input neuron with the id (2*threshold_counter + 1) spikes
    
    threshold_counter += 1 
        \end{lstlisting}

    \subsection{Energy and latency benchmarking results}
        The detailed benchmarking results for the ASP-SNN solving the sMNIST task on Loihi are given in Table~\ref{tab:sMNIST-power-results}. Here we compare the energy consumed, the latency (also called delay) to compute the output, and the resultant energy-delay product with an implementation of an LSTM running on a GPU.
        \clearpage

\begin{savenotes}
\begin{table*}[ht]
\setlength\tabcolsep{5pt}
\scaleboxratio{0.504141819262814335}{%
\begin{tabular}{c|c|c|ccc|c|c|c|c|c|c|c}
\textbf{}                                                                               & \textbf{}          & \textbf{}      & \multicolumn{3}{c}{\textbf{Power (mW)}}             & \textbf{Time per}       & \textbf{Latency}       & \textbf{Latency}       & \textbf{Energy per}     & \textbf{Energy}        & \textbf{Energy Delay}  & \textbf{EDP}           \\
\textbf{Hardware}                                                                       & \textbf{\# cores}  & \textbf{}      & \textbf{Static} & \textbf{Dynamic} & \textbf{Total} & \textbf{time step (µs)} & \textbf{(ms)}          & \textbf{ratio}         & \textbf{Inference (mJ)} & \textbf{ratio}         & \textbf{Product (µJs)} & \textbf{ratio}         \\ \hline
\multirow{3}{*}{\begin{tabular}[c]{@{}c@{}}Loihi\end{tabular}}        & \multirow{3}{*}{1} & x86 cores      & 0.08            & 24.33            & 24.41          & \multirow{3}{*}{16.79}  & \multirow{3}{*}{14.11} & \multirow{3}{*}{1.00x} & 0.34                    & \multirow{3}{*}{1.00x} & \multirow{3}{*}{5.14}  & \multirow{3}{*}{1.00x} \\
                                                                                        &                    & neuron cores   & 0.51            & 0.91             & 1.42           &                         &                        &                        & 0.02                    &                        &                        &                        \\
                                                                                        &                    & total          & 0.59            & 25.24            & 25.83          &                         &                        &                        & 0.36                    &                        &                        &                        \\ \hline
\multirow{3}{*}{\begin{tabular}[c]{@{}c@{}}Nvidia\\      RTX 2070\end{tabular}}         & \multirow{3}{*}{-} & batch size 1   & 33898.00        & 34602.00         & 68500.00       & -                       & 39.73                  & 2.82x                  & 2721.51                 & 7,467.20x              & 108125.39              & 21,025.64x             \\
                                                                                        &                    & batch size 50  & 33898.00        & 38171.00         & 72069.00       & -                       & 37.44                  & 2.65x                  & 53.97                   & 148.07x                & 2020.46                & 392.89x                \\
                                                                                        &                    & batch size 100 & 33898.00        & 53287.00         & 87185.00       & -                       & 40.41                  & 2.86x                  & 35.23                   & 96.67x                 & 1423.70                & 276.85x                \\ \hline
\multirow{3}{*}{\begin{tabular}[c]{@{}c@{}}Intel   Core\\       i5-7440HQ\end{tabular}} & \multirow{3}{*}{-} &                &                 &                  &                &                         &                        &                        &                         &                        &                        &                        \\
                                                                                        &                    & batch size 1   & 2040.00         & 18886.00         & 20926.00       & -                       & 83.15                  & 5.89x                  & 1740.00                 & 4,774.16x              & 144680.74              & 28,134.05x             \\
                                                                                        &                    &                &                 &                  &                &                         &                        &                        &                         &                        &                        &                       
\end{tabular}
}
\caption[Benchmark results for sMNIST]{\textbf{Benchmark results for sMNIST:}~Comparison of energy and time measurements of the spiking AHP-SNN network on Loihi against the corresponding LSTM on GPU and CPU solving the  sMNIST task\protect\footnotemark.}
\label{tab:sMNIST-power-results}
\end{table*}
\end{savenotes}

\footnotetext{Loihi: Nahuku board (ncl-ghrd-01), CPU: Intel Core i9-7920X, RAM: 128GB, OS: Ubuntu 16.04.6 LTS, NxSDK: 0.95\\
Nvidia RTX 2070: Nvidia RTX 2070 Super, GPU-RAM: 8GB, CPU: Intel Core i7-9700K, RAM: 32GB, OS: Ubuntu 16.04.6 LTS, Python 3.6.5, TensorFlow-GPU: 1.14.0, CUDA: 10.0.\\
Intel Core i5-7440HQ: RAM: 16GB, OS: Windows 10 (build18362), Python 3.6.7, TensorFlow: 1.14.1\\
Performance results are based on testing as of \today{} and may not reflect all publicly available security updates. Results may vary.}

\begin{figure}[h!]
    \centering
    \includegraphics[width=0.9\textwidth]{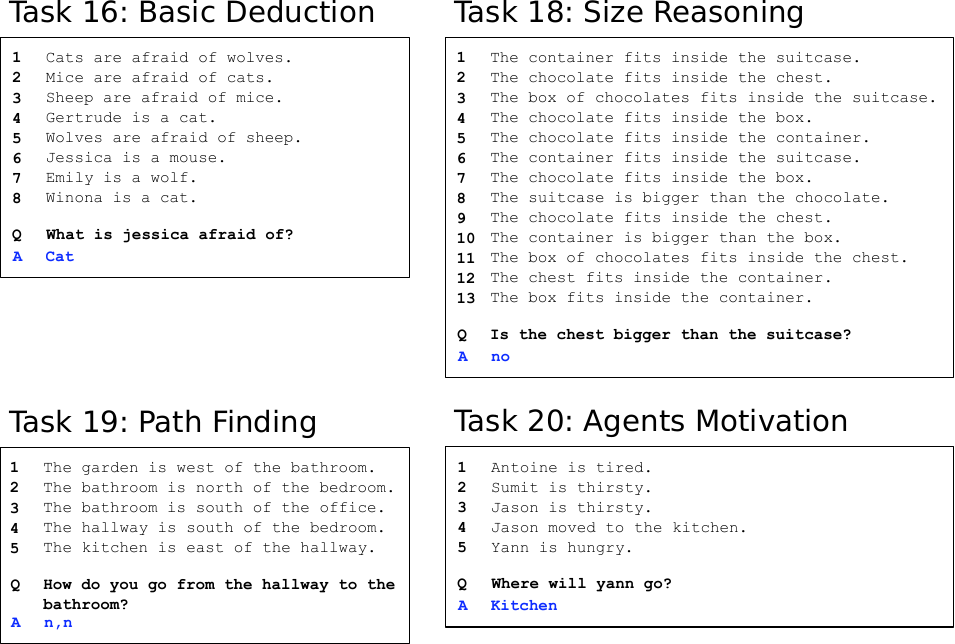}
    \caption[Examples for four types of tasks in the bAbI dataset]{\textbf{Examples for four types of tasks in the bAbI dataset:}~ 
             Each example consists of a story and a question which allows a single word answer (except for Task 19, where two directions can be given as answer).
             The dataset consists of 20 types of tasks, each targeting a different aspect of reasoning.
             In this figure, examples from four tasks are shown. Each one demonstrates the requirement for relational reasoning in order to successfully answer the question. ``Basic Deduction'', ``Path Finding'', and ``Size Reasoning'' require only reasoning across multiple sentences. Task 20  ``Agents Motivation'' requires in addition association of concepts with information that is external to the current story.}
    \label{fig:babi-examples}
\end{figure}
    
\section{Details for the Spiking RelNet on the bAbI task}

    \subsection{Output accuracy of the Spiking RelNet}
        
        The Spiking RelNet is trained on the combined data from 17 out of 20 bAbI tasks and it's performance is compared to an implementation of a non-spiking RelNet in Table~\ref{tab:bAbI-results}.
        We have excluded the 3 tasks "Task 2: Two Supporting Facts", "Task 3: Three Supporting Facts", and "Task 16: Basic Induction". For an example of some of the tasks on which the network was trained, see Fig.~\ref{fig:babi-examples}.
        This is because the non-spiking RelNet did not converge on these tasks.
        The network we trained was able to solve 16/17 tasks to within a 5\% Error, which is the threshold at which a task is considered solved (used in \cite{santoro2017}, and \cite{weston2015}).

        The task "Task 17: Positional Reasoning" has a rather high error, which we think is because the comparatively complex sentences in this task require a longer compute time to process.
        This is a trade-off we make for faster training and energy efficiency. With a larger number of time steps we find that Task 17 can in-fact be solved. To evidence this, we show two additional columns in Table~\ref{tab:bAbI-results}.

        The first column corresponds to a case where we simply extend the number of time steps for which we run the feed-forward part of the Spiking RelNet. Here we pad the embeddings upto a longer $\tsim = 45\ms$ (compared to $\tsim = 37\ms$ as specified in the Methods of the main text), and run the feed-forward part of the relational network (i.e. modules C--E in main text Fig. 4) for these many time steps. We observe here that while Task 17 is not yet under $5\%$ error, the error has dropped significantly.

        The second column shows a simulation where all the time constants, refractory periods, $\tinp$, $\tsim$, and time per word in embedding are tripled. The parameters are additionally adjusted so that the simulation is effectively slowed by a factor of 3. The resulting time lengths used are $\tinp=40\ms$, $\tsim=110\ms$, $\Tword=30\ms$. The tripled time constants are provided in table \ref{tab:neuron-params}.
        We see here that the additional temporal resolution allows all 17 tasks to be solved within $5\%$ classification error.
        However, with the 3-fold increase in time, we get a 4.5-fold (from 148.47 to 663.80) increase  in the energy-delay product. Roughly the same number of total spikes are used in both cases,

        This result demonstrates that it is possible to gain a significant gain in energy efficiency with a slight decrease in the overall performance (only a single task suffers loss in performance) in the bAbI task.

\begin{table}
\setlength\tabcolsep{5pt}
\scaleboxratio{0.751825118490240481}{
\begin{tabular}{l|cc|cc}
\hline 
\multirow{4}{*}{Task Name} & Spiking          & Non-spiking & \multicolumn{2}{c}{Spiking RelNet with }\tabularnewline
                           & RelNet           &  RelNet     & \multicolumn{2}{c}{increased compute time}\tabularnewline
\cline{4-5} \cline{5-5} 
                           & $\tsim=37$ steps &             & $\tsim=45$ steps & Stretched in time:\tabularnewline
                           &                  &             &                  & all times, time constants and\tabularnewline
                           &                  &             &                  & refractory period tripled\tabularnewline
\hline 
Task  1: Single Supporting Fact      & 1.0     & 0.4     & 0.6    & 1.0    \tabularnewline
Task  2: Two Supporting Facts        & --      & 20.8    & --     & --     \tabularnewline
Task  3: Three Supporting Facts      & --      & 25.0    & --     & --     \tabularnewline
Task  4: Two Argument Relations      & 0.1     & 0.0     & 0.0    & 0.1    \tabularnewline
Task  5: Three Argument Relations    & 2.3     & 0.6     & 1.2    & 2.3    \tabularnewline
Task  6: Yes/No Questions            & 0.2     & 0.0     & 0.3    & 0.4    \tabularnewline
Task  7: Counting                    & 0.7     & 0.6     & 0.6    & 1.4    \tabularnewline
Task  8: Lists/Sets                  & 0.5     & 0.1     & 0.3    & 0.9    \tabularnewline
Task  9: Simple Negation             & 0.1     & 0.1     & 0.1    & 0.7    \tabularnewline
Task 10: Indefinite Knowledge        & 2.3     & 1.8     & 1.5    & 1.7    \tabularnewline
Task 11: Basic Coreference           & 0.9     & 1.5     & 1.6    & 2.1    \tabularnewline
Task 12: Conjunction                 & 4.8     & 3.6     & 4.3    & 4.2    \tabularnewline
Task 13: Compound Coreference        & 3.9     & 2.5     & 3.6    & 3.6    \tabularnewline
Task 14: Time Reasoning              & 0.9     & 0.7     & 0.5    & 0.0    \tabularnewline
Task 15: Basic Deduction             & 0.1     & 0.0     & 0.0    & 0.0    \tabularnewline
Task 16: Basic Induction             & --      & 52.6    & --     & --     \tabularnewline
Task 17: Positional Reasoning        & 18.4    & 4.6     & 6.5    & 2.3    \tabularnewline
Task 18: Size Reasoning              & 1.5     & 0.6     & 0.8    & 0.2    \tabularnewline
Task 19: Path Finding                & 0.8     & 7.9     & 1.5    & 3.7    \tabularnewline
Task 20: Agent Motivations           & 0.0     & 0.0     & 0.0    & 0.4    \tabularnewline
\hline 
\end{tabular}
}
\caption[Percentage error of different architectures on the different tasks of the bAbI dataset]{
    \textbf{Percentage error of different architectures on the different tasks of the bAbI dataset: }~
    According to Weston et al.~\cite{weston2015}, tasks with errors under 5\% are considered solved.
    The four columns from left-to-right are as follows.
    \textbf{Column 1:}~ Error of the Spiking RelNet (version benchmarked in the main text) on the 17 tasks it was trained on. All but task 17 are solved to within 5\% Error.
    \textbf{Column 2:}~ Error of the non-spiking RelNet (our implementation) on the 20 tasks it was trained on.
    \textbf{Column 3:}~ Spiking RelNet where we increase just the number of compute steps of the feed-forward networks to $\tsim=45\ms$ (compared to $37\ms$ in the original network).
                        We see here an immediate improvement in the performance of task 17.
    \textbf{Column 4:}~ Spiking RelNet with all time lengths, refractory periods, and time constants tripled compared to the original network, and other parameters adjusted to effectively slow the entire simulation by a factor of three. We now have $\tsim=110\ms$, $\tinp=40\ms$, $\Tword=30\ms$ (compared to $\tsim=37\ms$, $\tinp=14\ms$, $\Tword=10\ms$ in the original network). Other parameters are compared in Table~\ref{tab:neuron-params}. With all times tripled, we can solve all 17 tasks to under 5\% error. However, when implemented, the EDP increases roughly 4.5-fold. Thus we see that we tradeoff accuracy on a single task for a significant improvement in energy efficiency}
\label{tab:bAbI-results}
\end{table}

    \subsection{Energy and latency benchmarking results for the Spiking RelNet}
        The detailed benchmarking results of the Spiking RelNet applied to the bAbI dataset are shown in Table~\ref{tab:power-results}, where we compare the Spiking RelNet on Loihi to a non-spiking RelNet on a GPU in terms of the energy consumed, the latency (also called delay) in the calculation of the output, and the energy-delay product. We also show there how this varies with task size.

\begin{savenotes}
\begin{table*}[ht]
\setlength\tabcolsep{5pt}
\scaleboxratio{0.496504002846579130}{%
\begin{tabular}{c|c|c|ccc|c|c|c|c|c|c|c}
\textbf{}                        & \textbf{\# sentences}                                                          & \textbf{}      & \multicolumn{3}{c|}{\textbf{Power (W)}}             & \textbf{Time per}       & \textbf{Latency}      & \textbf{Latency}       & \textbf{Energy per}     & \textbf{Energy}        & \textbf{Energy Delay}   & \textbf{EDP}                    \\
\textbf{Hardware}                & \textbf{\# cores}                                                              & \textbf{}      & \textbf{Static} & \textbf{Dynamic} & \textbf{Total} & \textbf{time step (µs)} & \textbf{(ms)}         & \textbf{ratio}         & \textbf{Inference (mJ)} & \textbf{ratio}         & \textbf{Product (µJs)}  & \textbf{ratio}                  \\ \hline
\multirow{3}{*}{Loihi}           & \multirow{3}{*}{\begin{tabular}[c]{@{}c@{}}20\\      2320 cores\end{tabular}}  & x86 cores      & 0.01            & 0.44             & 0.44           & \multirow{3}{*}{45.73}  & \multirow{3}{*}{6.54} & \multirow{3}{*}{1.00x} & 2.89                    & \multirow{3}{*}{1.00x} & \multirow{3}{*}{148.47} & \multirow{3}{*}{\textbf{1.00x}} \\
                                 &                                                                                & neuron cores   & 1.89            & 1.14             & 3.03           &                         &                       &                        & 19.82                   &                        &                         &                                 \\
                                 &                                                                                & total          & 1.90            & 1.57             & \textbf{3.47}  &                         &                       &                        & \textbf{22.70}          &                        &                         &                                 \\
\multirow{3}{*}{Nvidia RTX 2070} & \multirow{3}{*}{20}                                                            & batch size 1   & 33.50           & 5.89             & 39.39          & -                       & 2.51                  & 0.38x                  & 98.88                   & 4.36x                  & 248.18                  & \textbf{1.67x}                  \\
                                 &                                                                                & batch size 50  & 33.50           & 73.86            & 107.36         & -                       & 4.43                  & 0.68x                  & 9.51                    & 0.42x                  & 42.14                   & 0.28x                           \\
                                 &                                                                                & batch size 100 & 33.50           & 82.38            & 115.88         & -                       & 8.26                  & 1.26x                  & 9.57                    & 0.42x                  & 79.06                   & 0.53x                           \\ \hline
\multirow{3}{*}{Loihi}           & \multirow{3}{*}{\begin{tabular}[c]{@{}c@{}}16*\\      1552 cores\end{tabular}} & x86 cores      & 0.00            & 0.43             & 0.43           & \multirow{3}{*}{55.89}  & \multirow{3}{*}{7.99} & \multirow{3}{*}{1.00x} & 3.47                    & \multirow{3}{*}{1.00x} & \multirow{3}{*}{151.77} & \multirow{3}{*}{\textbf{1.00x}} \\
                                 &                                                                                & neuron cores   & 1.16            & 0.78             & 1.94           &                         &                       &                        & 15.52                   &                        &                         &                                 \\
                                 &                                                                                & total          & 1.17            & 1.21             & \textbf{2.38}  &                         &                       &                        & \textbf{18.99}          &                        &                         &                                 \\
\multirow{3}{*}{Nvidia RTX 2070} & \multirow{3}{*}{16}                                                            & batch size 1   & 33.36           & 5.47             & 38.82          & -                       & 2.6                   & 0.33x                  & 100.94                  & 5.32x                  & 262.45                  & \textbf{1.73x}                  \\
                                 &                                                                                & batch size 50  & 33.36           & 51.47            & 84.83          & -                       & 4.82                  & 0.60x                  & 8.18                    & 0.43x                  & 39.42                   & 0.26x                           \\
                                 &                                                                                & batch size 100 & 33.36           & 76.36            & 109.71         & -                       & 5.43                  & 0.68x                  & 5.96                    & 0.31x                  & 32.35                   & 0.21x                           \\ \hline
\multirow{3}{*}{Loihi}           & \multirow{3}{*}{\begin{tabular}[c]{@{}c@{}}10\\      700 cores\end{tabular}}   & x86 cores      & 0.01            & 0.44             & 0.45           & \multirow{3}{*}{36.36}  & \multirow{3}{*}{5.20} & \multirow{3}{*}{1.00x} & 2.33                    & \multirow{3}{*}{1.00x} & \multirow{3}{*}{58.73}  & \multirow{3}{*}{\textbf{1.00x}} \\
                                 &                                                                                & neuron cores   & 0.86            & 0.86             & 1.72           &                         &                       &                        & 8.96                    &                        &                         &                                 \\
                                 &                                                                                & total          & 0.87            & 1.30             & \textbf{2.17}  &                         &                       &                        & \textbf{11.30}          &                        &                         &                                 \\
\multirow{3}{*}{Nvidia RTX 2070} & \multirow{3}{*}{10}                                                            & batch size 1   & 33.90           & 4.63             & 38.53          & -                       & 2.28                  & 0.44x                  & 87.86                   & 7.78x                  & 200.31                  & \textbf{3.41x}                  \\
                                 &                                                                                & batch size 50  & 33.90           & 54.37            & 88.27          & -                       & 3.47                  & 0.67x                  & 6.13                    & 0.54x                  & 21.26                   & 0.36x                           \\
                                 &                                                                                & batch size 100 & 33.90           & 65.15            & 99.05          & -                       & 3.97                  & 0.76x                  & 3.93                    & 0.35x                  & 15.61                   & 0.27x                           \\ \hline
\multirow{3}{*}{Loihi}           & \multirow{3}{*}{\begin{tabular}[c]{@{}c@{}}6\\      332 cores\end{tabular}}    & x86 cores      & 0.01            & 0.45             & 0.46           & \multirow{3}{*}{27.64}  & \multirow{3}{*}{3.95} & \multirow{3}{*}{1.00x} & 1.81                    & \multirow{3}{*}{1.00x} & \multirow{3}{*}{29.11}  & \multirow{3}{*}{\textbf{1.00x}} \\
                                 &                                                                                & neuron cores   & 0.42            & 0.99             & 1.41           &                         &                       &                        & 5.56                    &                        &                         &                                 \\
                                 &                                                                                & total          & 0.43            & 1.44             & \textbf{1.86}  &                         &                       &                        & \textbf{7.37}           &                        &                         &                                 \\
\multirow{3}{*}{Nvidia RTX 2070} & \multirow{3}{*}{6}                                                             & batch size 1   & 33.80           & 5.58             & 39.38          & -                       & 2.23                  & 0.56x                  & 87.82                   & 11.92x                 & 195.84                  & \textbf{6.73x}                  \\
                                 &                                                                                & batch size 50  & 33.80           & 44.76            & 78.56          & -                       & 3.2                   & 0.81x                  & 5.03                    & 0.68x                  & 16.09                   & 0.55x                           \\
                                 &                                                                                & batch size 100 & 33.80           & 52.82            & 86.62          & -                       & 3.76                  & 0.95x                  & 3.26                    & 0.44x                  & 12.25                   & 0.42x                           \\ \hline
\multirow{3}{*}{Loihi}           & \multirow{3}{*}{\begin{tabular}[c]{@{}c@{}}2\\      124 cores\end{tabular}}    & x86 cores      & 0.01            & 0.46             & 0.47           & \multirow{3}{*}{22.96}  & \multirow{3}{*}{3.28} & \multirow{3}{*}{1.00x} & 1.53                    & \multirow{3}{*}{1.00x} & \multirow{3}{*}{18.36}  & \multirow{3}{*}{\textbf{1.00x}} \\
                                 &                                                                                & neuron cores   & 0.17            & 1.07             & 1.24           &                         &                       &                        & 4.06                    &                        &                         &                                 \\
                                 &                                                                                & total          & 0.18            & 1.53             & \textbf{1.70}  &                         &                       &                        & \textbf{5.59}           &                        &                         &                                 \\
\multirow{3}{*}{Nvidia RTX 2070} & \multirow{3}{*}{2}                                                             & batch size 1   & 33.27           & 4.99             & 38.26          & -                       & 2.41                  & 0.73x                  & 92.20                   & 16.49x                 & 222.21                  & \textbf{12.10x}                 \\
                                 &                                                                                & batch size 50  & 33.27           & 41.62            & 74.89          & -                       & 2.92                  & 0.89x                  & 4.37                    & 0.78x                  & 12.77                   & 0.70x                           \\
                                 &                                                                                & batch size 100 & 33.27           & 46.38            & 79.64          & -                       & 3.48                  & 1.06x                  & 2.77                    & 0.50x                  & 9.64                    & 0.53x                          
\end{tabular}
}
\caption[Benchmark results for question-answering task using RelNet.]{\textbf{Benchmark results for question-answering task using RelNet}~Benchmarking comparison and scaling analysis of the Spiking RelNet on Loihi against the corresponding ANN on GPU\protect\footnotemark. The data set was grouped by number sentences per sample which in turn determines the number of AHP-SNNs and therefore cores per sample. Measurements were done using 250 input samples. The energy per inference was calculated using total power values.\\
*For network size 16 only 100 input samples were used, as there are not enough test samples containing 16 sentences.}
\label{tab:power-results}
\end{table*}
\end{savenotes}

\footnotetext{Loihi: Nahuku board (ncl-ghrd-01), CPU: Intel Core i9-7920X, RAM: 128GB, OS: Ubuntu 16.04.6 LTS, NxSDK: 0.95\\
GPU: Nvidia RTX 2070 Super, GPU-RAM: 8GB, CPU: Intel Core i7-9700K, RAM: 32GB, OS: Ubuntu 16.04.6 LTS, Python 3.6.5, TensorFlow-GPU: 1.14.0, CUDA: 10.0.\\ Performance results are based on testing as of \today{} and may not reflect all publicly available security updates. Results may vary.}
        \clearpage

    \subsection{Details of the Spiking RelNet architecture}
        \subsubsection*{The embedding networks}

            In order to embed sentences and questions into spike sequence, we use AHP-SNN Networks. The AHP-SNN network for sentences uses a different set of weights than those used by the AHP-SNN network for questions, however all other parameters are identical and are described below.

            Each AHP-SNN contains $100$ LIF-neurons and $100$ AHP-neurons.
            The synaptic delays take an integer value uniformly from 1 to 3 ms.
            Note here that all time lengths and time constants are specified in terms of computation steps on Loihi, with 1 computation step corresponding to $1\ms$.

            The input to the AHP-SNN is described here. We assign each distinct word from the bAbl dataset an index, and associate an input neuron.
            When a word is provided as input to the AHP-SNN, the corresponding input neuron, and only that neuron, emits spikes continuously for $\Tword=10\ms$.
            To present a sentence or question to the AHP-SNN, each word in it is encoded as above, and input to the AHP-SNN in sequence so that the first word is aligned to the final time step.
            Thus the input to the AHP-SNN takes at most $10\ms*\Nwords = 110\ms$ where $\Nwords = 11$ is the maximum number of words in a sentence or question of the bAbI task.
            These input neurons are connected to the AHP-SNN in an all-to-all manner.

            The final embedded spike sequences $o_i(t)$/$q(t)$ are formed by taking The spike activity of the AHP-SNN over the last $\tinp = 14\ms$, and padding them with zeros up to a total compute time of $\tsim = 37\ms$.
            These embeddings are the input to the instances of the relational function $g_\theta(o_i(t), o_j(t), q(t))$, and consequently all the feed-forward LIF-neuron networks from this point on are run for a length of $\tsim$.
            The zero padding and the longer compute time $\tsim > \tinp$ are necessary so that the spikes from the embedding have enough time steps to propagate through the several layers of the feed-forward LIF-neuron networks.


        \subsubsection*{The relational function}

            Currently, the relational function $g_\theta$ is implemented as a 4-layer feed-forward network of LIF-neurons without AHP-currents, with 256 neurons in each layer.
            Each layer is connected all-to-all to the next. The synaptic delays take up values from 1 to 3 ms.
            The relational function takes as input a triplet containing a pair of embedded sentences and the embedded question $\robrace{o_i(t), o_j(t), q(t)}$, and returns an output spike sequence.
            The spike sequences of the sentence pair and question are weighted and fed via all-to-all connections to the first layer.

            The above relational function is applied once for each pair of sentences $i, j$ in the story where $i \le j$. Note that there are instances of $g_\theta$ that receive the same sentence twice i.e. $i = j$
            The output of each instance of the relational function is a spike matrix,

        \subsubsection*{The aggregation layer}

            The aggregation layer is an addition to the original architecture proposed in Santoro et. al.\cite{santoro2017}, which is necessary due to constraints of neuromorphic hardware. These constraints are discussed in section \ref{sub:loihi-connectivity-constraints}.
            The aggregation is a single layer of LIF-neurons without AHP-currents, such that the output of each instance of $g_\theta$ is connected one-to-one to this layer.
            This implements an element-wise function $f_{agg}$ on the outputs of $g_\theta$ instances summed.
            Each neuron of the aggregation layer receives as input the sum of the output spikes across all $g_\theta$ instances, and outputs a spike sequence.

        \subsubsection*{The readout function}

            The readout function $f_\phi(\cdot)$ is implemented using a 3 layer feed-forward LIF-neuron network, with layer sizes 256, 512, 160,
            followed by a specially designed linear readout that is described in the methods section of the main text.
            Each layer, including the readout, is connected all-to-all to the next.
            The synaptic delays take up values from 1 to 3 ms.

    \subsection{Details of Spiking RelNet training}

        \subsubsection*{Pretraining the AHP-SNN networks}

            In order to reduce the number of epochs required to train the Spiking RelNet, we choose to pre-train the AHP-SNN's that embed the sentences and questions as below.
            We first train a non-spiking implementation of the RelNet end-to-end until we reach optimal performance.
            The non-spiking AHP-SNN uses LSTM units to embed the sentences and questions into representations.
            We then train the AHP-SNN to reproduce the output of these pre-trained LSTM networks for all the sentences used in the database.

            In order to readout a value from the AHP-SNN, which can be compared to the LSTM output, we use a linear readout similar to the one used to train the entire Spiking RelNet (described in the main Methods).
            The number of readout units matches the dimension of the LSTM that we wish to approximate i.e. 32.
            We then compare the value of this readout to the output of the LSTM, using mean squared error.
            This mean squared error is used as the loss function to train the AHP-SNN weights using back-propagation in time (BPTT).

            The weights of the AHP-SNN are then frozen, and the resultant spike embeddings learnt here are used as the input when training the weights of the feed-forward part of the Spiking RelNet.
            Note that when we use the spikes of these pre-trained AHP-SNN's as input to the $g_\theta$ instances, the readout weights that were used to pre-train the AHP-SNN are not used in any way.
            We directly connect the AHP-SNN spikes to $g_\theta$ using randomly initialized weights and train these weights.
            We find that, once pre-trained, all the information pertaining to the sentence is encoded within the spikes of the AHP-SNN, which can then be processed by the feed-forward part.

        \subsubsection*{Rate regularization for $g_\theta$ instances}
            For all other layers, the rate regularization pushes the mean rate of each neuron across the batch toward a specific target rate.
            However we use a more aggressive regularization for the spike rates of the instances of $g_\theta$.

            Corresponding to each layer of $g_\theta$, we calculate the following loss.
            Consider the $b$'th story in the batch.
            For this story, we denote spike rate of the neuron $k$, in the $(i, j)$'th instance of $g_\theta$ as $\rho^b_{k,ij}$
            Now we define $R^b_k = \sum_{1\le i \le j \le M} \rho^b_{k, ij}$ as the total spike rate of neuron $k$ across all $g_\theta$ instances for the $b$'th story.
            We then define the rate regularization loss as below:
            \begin{equation*}
                L_R = \lambda_R (\mean_k\;(\mean_b(R^b_k) - R_{\,\text{target}})^2)^2.
            \end{equation*}

            We calculate a similar loss for each layer of $g_\theta$ and sum them up as a part of the final loss.
            For each neuron, this regularization loss pushes the sum of its spike rate across the $g_\theta$ instances to a target value of $R_\text{target}$.
            For our networks, $R_{\,\text{target}} = 300\Hz$. For a task with two sentences, this translates to a target of 3.7 spikes per neuron, and for a task with 20 sentences, this corresponds to 0.05 spikes per neuron. The corresponding spike rates achieved when trained can be seen in the main text Fig. 3 B.

    \subsection{Constraints on connectivity on Loihi}
        \label{sub:loihi-connectivity-constraints}
        In this section we discuss the restrictions on network connectivity and the strategies used to place a Spiking RelNet that processes stories up to $M=20$ sentences in length.
        In order to understand the following section, it is useful to define some of the relevant terminology:
        \begin{description}
            \item[Neuro-Core]
                A neuro-core (short for neuromorphic core) is a fundamental computational element in the Loihi chip. Each neuro-core can compute the dynamics of up to 1024 Neurons, and contains a shared SRAM which contains data pertaining to the weights of the incoming synapses as well as shared configuration and state variables.

            \item[Chip]
                A Loihi chip is a block of 128 interconnected neuro-cores integrated within the same silicon substrate. It is possible for multiple chips to be connected together to allow for a larger number of interconnected neuro-cores and thereby larger networks.

            \item[Axon]
                The axon is a structure that is part of the infrastructure that implements connectivity and spike transport in Loihi. Loihi implements the connectivity between different neuro-cores via inter-core connections called axons. Each axon indexed by $(i, C)$ is a connection between a presynaptic neuron $i$ and a postsynaptic neuro-core $C$. If an axon $(i, C)$ is connected, all spikes generated by the presynaptic neuron $i$, are routed through this axon to neuro-core $C$, where it is weighted by the relevant weights and delivered to the postsynaptic neurons. Each axon $(i, C)$ is considered to be an \emph{input axon} of neuro-core $C$ and an \emph{output axon} of the neuro-core that contains neuron $i$.
        \end{description}

        We now discuss here the various constraints and their impact on the network architecture and placement.
        
         \subsubsection*{Synaptic memory limit}
            As mentioned earlier, the SRAM in a neuro-core is used to store the parameters for any incoming connection to that neuro-core.
            The limited per-neuro-core SRAM memory for synaptic parameters puts a limit on the number of incoming synapses to a particular neuro-core.
            Except the aggregation layer, all layers in the network have densely connected input synapses which translates to a larger number of incoming connections than can fit into a single neuro-core.
            Thus, need to appropriately split the postsynaptic neurons across several neuro-cores so that the total number of synapses coming into each neuro-core can fit into the memory.

        \subsubsection*{Limits pertaining to fanouts -- AHP-SNN relay layer}
            Loihi has two limits that pertain to the fanout of a layer.

            \begin{itemize}
                \item \emph{output axon limit} -- The number of outgoing axons from a neuro-core is limited in general to 2048,
                      and 4096 if all the outgoing axons are connected to neuro-cores within the same chip.
                \item \emph{neuro-core fanout limit} -- The number of different neuro-cores to which a single neuron can be connected to is at-most 512.
            \end{itemize}

            This constraint plays a role only in the case of the connections from the AHP-SNN's to the first layer of the instances of $g_\theta$.
            To see this, we have $\frac{M(M+1)}{2} = 210$ instances of $g_\theta$.
            For a particular sentence $k$, there are exactly $M$ pairs $(i, j) :: 1\le i \le j\le M$ that contain $k$.
            There is also the pair $(k, k)$ which contains $k$ twice.
            This means that the output of the corresponding AHP-SNN-$k$ ($o_k(t)$) is connected to a $g_\theta$ instance $M+1$ times in an all-to-all manner.
            It takes $4$ neuro-cores (Table~\ref{tab:n-cores-per-layer}) to place the first layer of an instance of $g_\theta$.
            This implies that each neuron of AHP-SNN-$k$ is connected to $4\times(M+1) = 84$ neuro-cores, implying $84$ output axons per AHP-SNN neuron.
            Similarly for the question-AHP-SNN, since it forms an input to all the instances of $g_\theta$, we get $4\times\frac{M(M+1)}{2}\ = 840$ output axons per AHP-SNN neuron.

            Given the above number of output axons per neuron, the output axon limit puts a stronger limit on the number of neurons per neuro-core than the synaptic memory limit.
            However, the AHP-SNN is a recurrent network and inter-core communication will lead to significant latency if the neurons of the AHP-SNN are spread across too many neuro-cores.
            Moreover, for the question-AHP-SNN, that fact that each neuron fans out to $840$ neuro-cores means that we violate the neuro-core fanout limit.

            This motivates the use of relay networks.
            A \emph{relay layer}, as the name suggests relays spikes from the layer that forms the input.
            It has the same number of neurons as the input layer and the input layer is connected in a one-to-one manner to it.
            Each time a neuron of the relay layer receives a spike, it generates a spike, thereby reproducing the input spike train as the output spike train.

            Each AHP-SNN is connected to multiple relay layers, each of which fans-out to a subset of the $g_\theta$ instances that take the corresponding sentence/question as input.
            Since the connection from the AHP-SNN to the relay neurons is one-to-one, and we fanout to fewer relay networks than the original fanout to the $g_\theta$ instances, we have a much reduced fanout from the AHP-SNN. So much so that this is no longer the a constraint in placing the AHP-SNN.
            Also each relay neuron fans out to fewer neuro-cores than the original AHP-SNN, and can be split across neuro-cores without additional latency cost, thus satisfying all fanout constraints.

            The actual choice of relays and the fanouts from the relays is determined in a manner that minimize congestion in spike transport (described below in section \ref{sub:optimized-network-placement}).

            Each sentence-AHP-SNN fans out to 4 relay networks and the question-AHP-SNN fans out to 10.
            Each AHP-SNN can be placed on 2 neuro-cores (see Table~\ref{tab:n-cores-per-layer}), meaning each neuro-core has 100 neurons.
            The number of output axons per neuro-core for the AHP-SNN's are then $4*100 = 400$ and $10*100 = 1000$ for the sentence-AHP-SNN and question-AHP-SNN respectively, both well within the output axon limits.

        \subsubsection*{Input axon limit -- the aggregation layer}
            Loihi limits each neuro-core to have a maximum of $4096$ input axons.
            Note here that if a neuron $i$ is connected to even a single neuron in neuro-core $C$, the axon $(i, C)$ is an input axon of neuro-core $C$.
            In this case, we denote neuron $i$ as being presynaptic to the neuro-core $C$.
            The input axon limit means that for any neuro-core, a maximum of $4096$ neurons can be presynaptic to that neuro-core.
            Unlike the above two constraints which can be worked around,
            the input axon limit introduces a fundamental restriction to the network architectures that can be implemented on Loihi.

            In the original formulation of relational networks in \cite{santoro2017}, the outputs from all the relational function instances are summed together to form the input to the final readout function $f_\phi$.
            In an ideal scenario, it is possible to implement this using spiking networks using the fact that incoming spike inputs can be summed into the PSC's of the postsynaptic neurons.
            However, when the summed spike train forms an input that is connected all-to-all into the $f_\phi$ network, it becomes a problem.
            To implement this, we would need to connect each relational function instance in an all-to-all manner to the final readout function, with shared weights across the different instances.
            However, consider that the output dimension of the relational function is 256.
            This connectivity would mean that if we consider a single neuron of the first layer of $f_\phi$, this neuron would receive input from all $256\cdot \frac{M(M+1)}{2} = 53760$ output neurons across the instances of $g_\theta$.
            This means that a neuro-core with even a single such neuron would have $53760$ presynaptic neurons, and thus input axons, which is completely beyond the input axon limit.

            In order to deal with this constraint, we have modified the original architecture of the Relational Network by adding an \emph{aggregation layer}.
            The aggregation layer is a layer of spiking LIF-neurons, that has the same number of neurons as the output layer of $g_\theta$, and to which each instance of $g_\theta$ is connected one-to-one with weights shared across the $g_\theta$ instances.
            This leads to each neuron from the aggregation layer having $210$ presynaptic neurons.
            The input axon limit thus allows up to $19$ neurons per neuro-core, and the $256$ neurons of the aggregation layer can be placed onto $14$ neuro-cores.
        
        In Table~\ref{tab:n-cores-per-layer}, we give the number of neuro-cores required to place an instance of a layer is given for the different layers of the Spiking RelNet.
        Also mentioned are the constraints that lead to the layers being split across as many neuro-cores.

\begin{table}[tb]
    \centering
    \setlength\tabcolsep{5pt}
    \scaleboxratio{0.799355570217131100}{
    \begin{tabular}{llllll}
        \hline
        \hline
            Relational Network         &            & Number of cores &                    & Number of                &               \\ [-0.7ex]
            Layer                      & Layer Size & per instance    & Connectivity limit & Instances                & Total Cores   \\
            \hline
            AHP-SNN (Sentences)           & 200        & 2               & Synaptic Memory    & $M=20$                   & 40            \\
            AHP-SNN (Question)            & 200        & 2               & Synaptic Memory    & 1                        & 2             \\
            Relay networks (sentences) & 200        & 1-2*            & Output Axon        & $4M=80$*                 & 100           \\
            Relay networks (questions) & 200        & 3-5*            & Output Axon        & 10*                      & 42            \\
            $g_\theta$ Layer 1         & 256        & 4               & Synaptic Memory    & $\frac{M(M+1)}{2} = 210$ & 840           \\
            $g_\theta$ Layer 2         & 256        & 2               & Synaptic Memory    & 210                      & 420           \\
            $g_\theta$ Layer 3         & 256        & 2               & Synaptic Memory    & 210                      & 420           \\
            $g_\theta$ Layer 4         & 256        & 2               & Synaptic Memory    & 210                      & 420           \\
            Aggregation Layer          & 256        & 14              & Input Axon         & 1                        & 14            \\
            $f_\phi$ Layer 1           & 256        & 2               & Synaptic Memory    & 1                        & 2             \\
            $f_\phi$ Layer 2           & 512        & 4               & Synaptic Memory    & 1                        & 4             \\
            $f_\phi$ Layer 3           & 160        & 3               & Synaptic Memory    & 1                        & 3             \\
            Linear Readout Layer       & 180        & 1               & Synaptic Memory    & 1                        & 1             \\
            \hline
                                       &            &                 &                    & \multicolumn{1}{r}{Total Cores: } & 2308 \\
            \hline
            \multicolumn{4}{l}{\small * Determined by placement scheme that minimizes spike congestion.} & & \\
    \end{tabular}
    }
    \caption[Parameters for the placement of the Spiking RelNet onto Loihi]{\textbf{Parameters for the placement of the Spiking RelNet onto Loihi:}~This table gives, for each layer of the relational network, the number of cores required to place all the neurons in a single instance of that layer. Among the three limits on connectivity, i.e. limits on synaptic memory, input axons, and output axons, the limit that ultimately results in the layer needing to be divided among multiple cores is specified as the corresponding connectivity limit. Also mentioned are the number of instances of each network corresponding to a maximum story size of $M=20$ sentences, and the total number of cores needed.}
    \label{tab:n-cores-per-layer}
\end{table}

    \subsection{Optimized network placement to minimize spike congestion}
    \label{sub:optimized-network-placement}
        In the methods section of the main text, we discuss how it is necessary to place the relay layers and the the several instances of the relational function $g_\theta$ in a careful manner so that the amount of cross-chip communication is minimized.

        In order to perform this optimization, it is quite difficult to actually optimize the cross-chip communication as that is not easy to calculate directly as a function of the network placement.
        Instead, we first notice that a majority of the spike congestion occurs when transferring the spikes from the AHP-SNN networks, via the relay layers to the input of the various instances of $g_\theta$.
        Thus it makes sense to optimize the placement of the relay layers and first layer of the instances $g_\theta$ in such a manner that the number of connections across chips is minimized.
        This objective is translated into the following constraints on the network placement.
        \begin{itemize}
            \item Firstly, a relay layer that forms an input to an instance of $g_\theta$ is located on the same chip as the first layer of that instance. That is, we place initial layer of the $g_\theta$ instances on the same chip as the relay layers that give them their corresponding input sentence and question embeddings.

            \item With this constraint, the connections from the AHP-SNN's to the relay layers become the cross-chip communication that needs to be optimized, meaning that we need to minimize the number of relay layers used.
                  Each chip has a limit of $128$ neuro-cores and thus a limit on the number of $g_\theta$ instances whose initial layers can be placed on a single chip.
                  Thus, when choosing the $g_\theta$ instances to be placed on the same chip, we select them such that the number of distinct sentences needed as input for this set of instances is minimized, thus minimizing the number of relay layers.
        \end{itemize}

\begin{figure}[h]
            \centering
            \includegraphics[width=\textwidth]{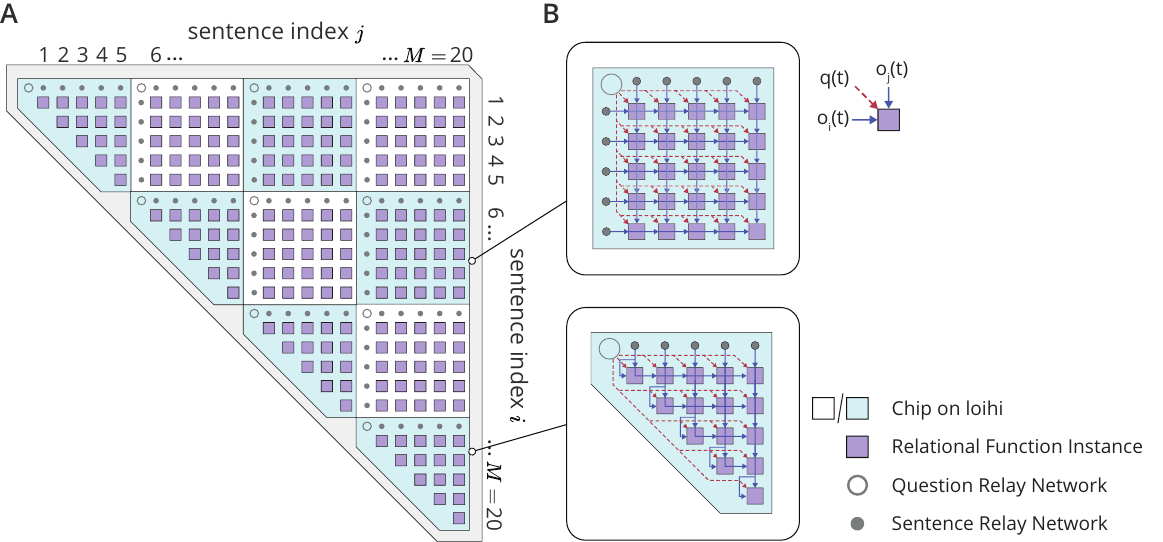}
            \caption[Placement of relay networks and the initial layer of $g_\theta$ instances that minimizes spike congestion.]{
                     \textbf{Placement of relay networks and the initial layer of $g_\theta$ instances that minimizes spike congestion: }~
                     \textbf{A)}~Here we show how the first layer of the several instances of the relational function $g_\theta$ are placed across the several chips.
                     The instances of $g_\theta$ are arranged according to the indices of the input sentences $o_i(t), o_j(t) : i \le j$.
                     Each blue/cyan block corresponds to a single Loihi chip, within which we show the $g_\theta$ instances whose first layer is placed on that chip, along with the relay layers that give them the input.
                     Each cell also has a relay layers for the question embedding $q(t)$.
                     The instances are grouped into squares because this minimizes the number of distinct inputs (i.e. relay layers) needed.
                     \textbf{B)}~A blowup of a chip showing the connections between the relay layers and the contained instances of $g_\theta$.}
            \label{fig:optimized-placement}
        \end{figure}

        The resultant placement of the relay networks and the initial layer of the instances of $g_\theta$ is detailed in Fig.~\ref{fig:optimized-placement}.

        For the subsequent layers of the $g_\theta$ instances, we simply place each instance in sequence with only the constraint that all the remaining 3 Layers should lie on the same chip.
        We find that the cross-chip communication from the first layer to the subsequent layers is very minimal and does not cause a significant delay.

        The advantage of such this layout is two-fold.
        Firstly, since a majority of the connections are within the chip, this significantly reduces the congestion that happens when transporting the spikes across chips.
        Secondly, since each relay network fans out only to neuro-cores that are placed on the same chip, the output axon limit is now $4096$ rather than $2048$ thus requiring fewer relay networks.

\clearpage

\bibliographystyle{apalike}
\bibliography{./bib/bib}